\documentclass[11pt]{article}

\usepackage[final]{acl}

\usepackage{times}
\usepackage{latexsym}

\usepackage[T1]{fontenc}

\usepackage[utf8]{inputenc}

\usepackage{microtype}

\usepackage{inconsolata}

\usepackage{graphicx}

\usepackage{amsmath, amssymb}
\usepackage{bbm}

\usepackage{booktabs}
\usepackage{array}
\usepackage{tabularx}
\usepackage[table]{xcolor}   
\usepackage{subcaption}      
\usepackage{arydshln}        
\usepackage{graphicx}        

\usepackage{algorithm}
\usepackage[noend]{algpseudocode}

\usepackage{enumitem}
\usepackage{wrapfig}
\usepackage{listings}
\usepackage{float}

\usepackage{tcolorbox}
\tcbuselibrary{breakable}

\definecolor{softGreen}{HTML}{D5E8D4}
\definecolor{softPink}{HTML}{F8D7DA}
\definecolor{lightpurple}{RGB}{235,225,250}
\definecolor{lemon}{RGB}{255, 250, 160}

\newcommand{\vcenteredcell}[1]{\raisebox{-.5\height}{#1}}

\algrenewcommand\algorithmicrequire{\textbf{Input:}}
\algrenewcommand\algorithmicensure{\textbf{Output:}}
\newcommand{\Phase}[1]{\State \textcolor{green!50!black}{\textit{// #1}}}

\usepackage{pifont}
\newcommand{\cmark}{\textcolor{green!60!black}{\ding{51}}}
\newcommand{\xmark}{\textcolor{red}{\ding{55}}}
\newcommand{\uparroweff}{\textcolor{green!60!black}{$\uparrow$}}
\newcommand{\downarroweff}{\textcolor{red}{$\downarrow$}}

\newcommand{\zsfootnote}{\footnote{\label{fn:zs_def}%
  \hangindent=1.5em \hangafter=1
  MMLU STEM is 5-shot; all others are zero-shot.}}
\newcommand{\zsref}{\textsuperscript{\ref{fn:zs_def}}}

\title{LightReasoner: Can Small Language Models Teach Large Language Models Reasoning?}
\def\model{LightReasoner}

\author{
  Jingyuan Wang\textsuperscript{1}\thanks{Equal contribution.} \quad
  Yankai Chen\textsuperscript{2}\footnotemark[1] \quad
  Zhonghang Li\textsuperscript{1} \quad
  Chao Huang\textsuperscript{1}\thanks{Corresponding author.} \\
  \textsuperscript{1}University of Hong Kong \quad
  \textsuperscript{2}University of Chicago \\
  \texttt{\{jameswangjingyuan,xiaoshu0712,bjdwh.zzh,chaohuang75\}@gmail.com}
}

\begin{document}
\maketitle

\begin{abstract}
Large language models (LLMs) have demonstrated remarkable progress in reasoning, often through supervised fine-tuning (SFT). However, SFT is resource-intensive, relying on large curated datasets, rejection-sampled demonstrations, and uniform optimization across all tokens—even though only a fraction carry meaningful learning value. In this work, we explore a counterintuitive idea: can smaller language models (SLMs) teach larger language models (LLMs) by revealing high-value reasoning moments that reflect the latter's unique strength? We propose \textbf{\model}\footnote{\textcolor{blue}{\url{https://github.com/HKUDS/LightReasoner}}}, a novel framework that leverages the behavioral divergence between a stronger \emph{expert} model (LLM) and a weaker \emph{amateur} model (SLM). \model\ operates in two stages: (1) a \emph{sampling} stage that pinpoints critical reasoning moments and constructs supervision examples capturing the expert's advantage through expert–amateur contrast, and (2) a \emph{fine-tuning} stage that aligns the expert model with these distilled examples, amplifying its reasoning strengths. Across seven benchmarks, \model\ improves accuracy by up to 28.1\%, while reducing time consumption by 90\%, sampled problems by 80\%, and tuned token usage by 99\%, all without relying on ground-truth labels. By turning weaker SLMs into effective teaching signals, \model\ offers a scalable and resource-efficient approach for advancing LLM reasoning.

\end{abstract}

\section{Introduction}
Large language models (LLMs) have achieved remarkable progress in language understanding and generation \citep{kaplan2020scaling, touvron2023llama}. However, they continue to struggle with systematic reasoning tasks that demand step-by-step logical precision. A prevalent strategy to address this gap is to align models with demonstrations of correct problem-solving trajectories. To this end, recent works \citep{yang2024qwen2, guo2025deepseek} often integrate rejection sampling \citep{yuan2023scalingrelationshiplearningmathematical} into supervised fine-tuning (SFT). While effective, this strategy is resource-intensive: it requires generating multiple candidate solutions, filtering them against ground-truth answers to retain only correct trajectories, and then fine-tuning on every token of those trajectories, treating trivial and crucial reasoning steps as equally valuable.

\begin{figure}[t!]  
    \centering

    \begin{subfigure}[t]{0.5\linewidth}
        \centering
        \includegraphics[width=\linewidth]{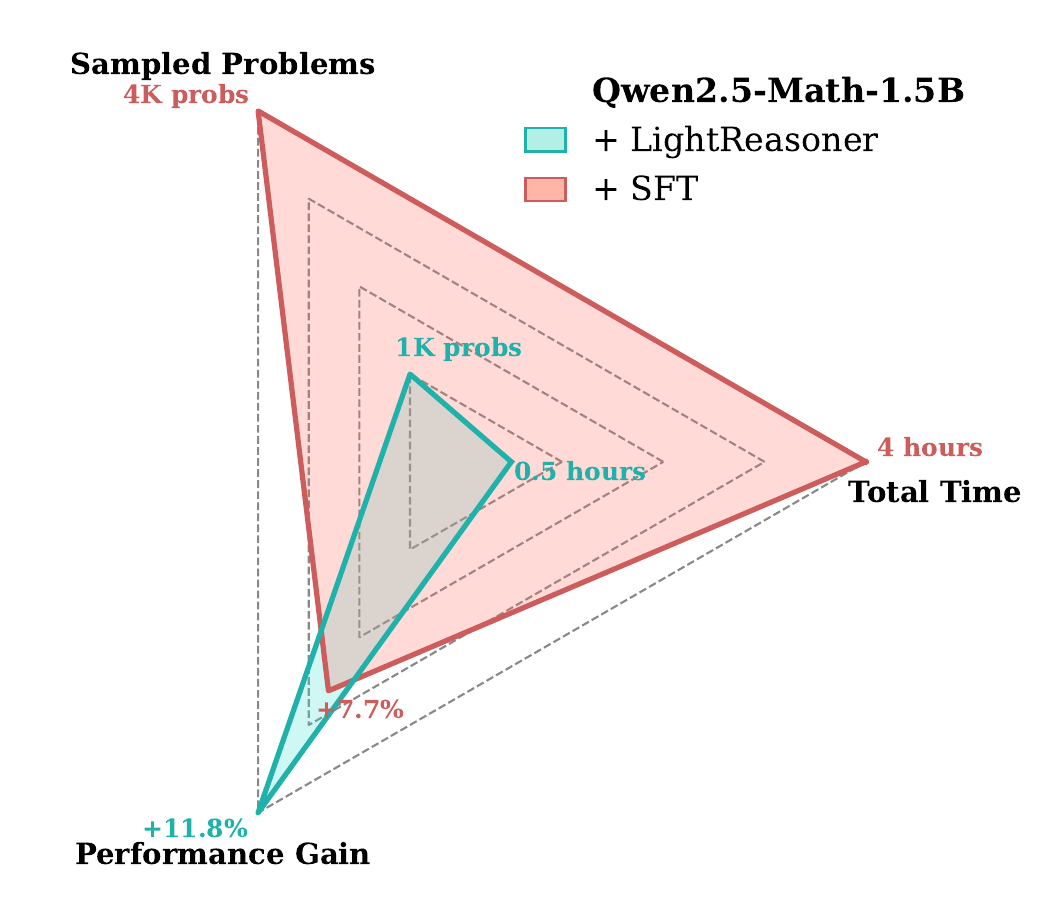}

        \scriptsize \textbf{(a)} Qwen2.5-Math-1.5B
    \end{subfigure}\hfill
    \begin{subfigure}[t]{0.5\linewidth}
        \centering
        \includegraphics[width=\linewidth]{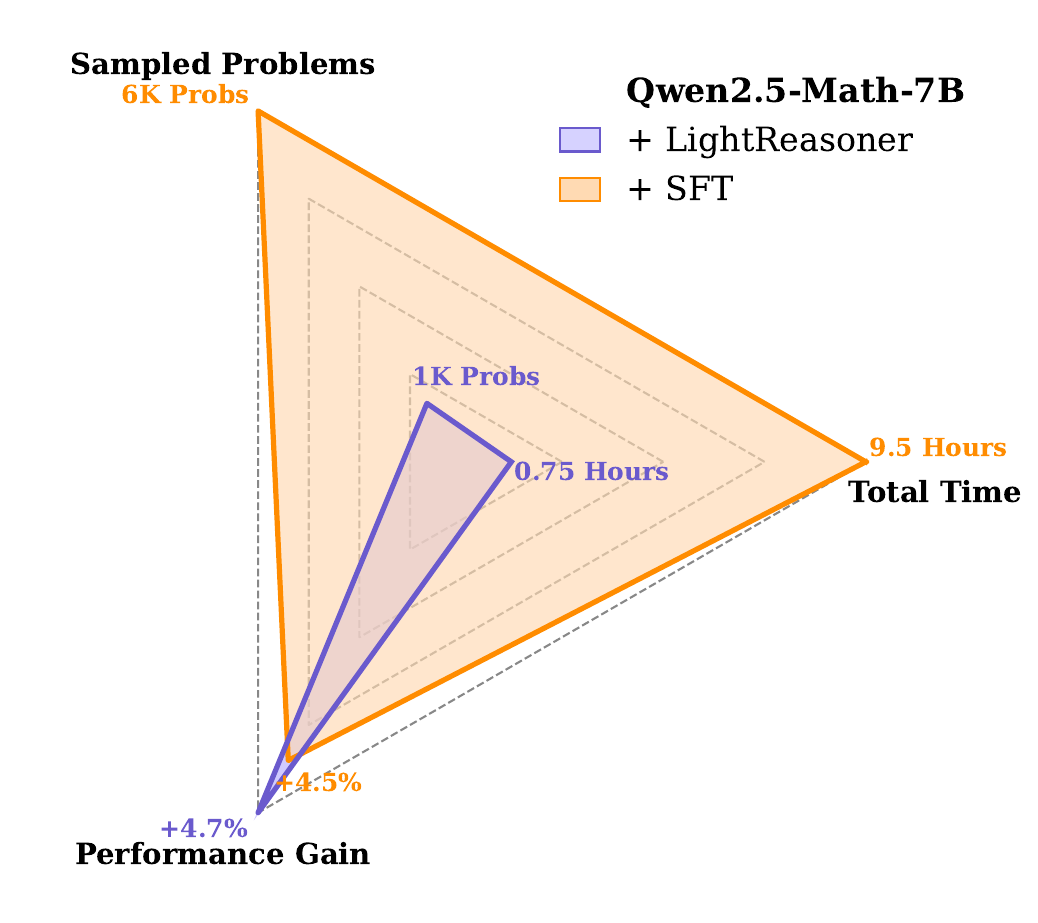}

        \scriptsize \textbf{(b)} Qwen2.5-Math-7B
    \end{subfigure}

    \begin{subfigure}[t]{0.5\linewidth}
        \centering
        \includegraphics[width=\linewidth]{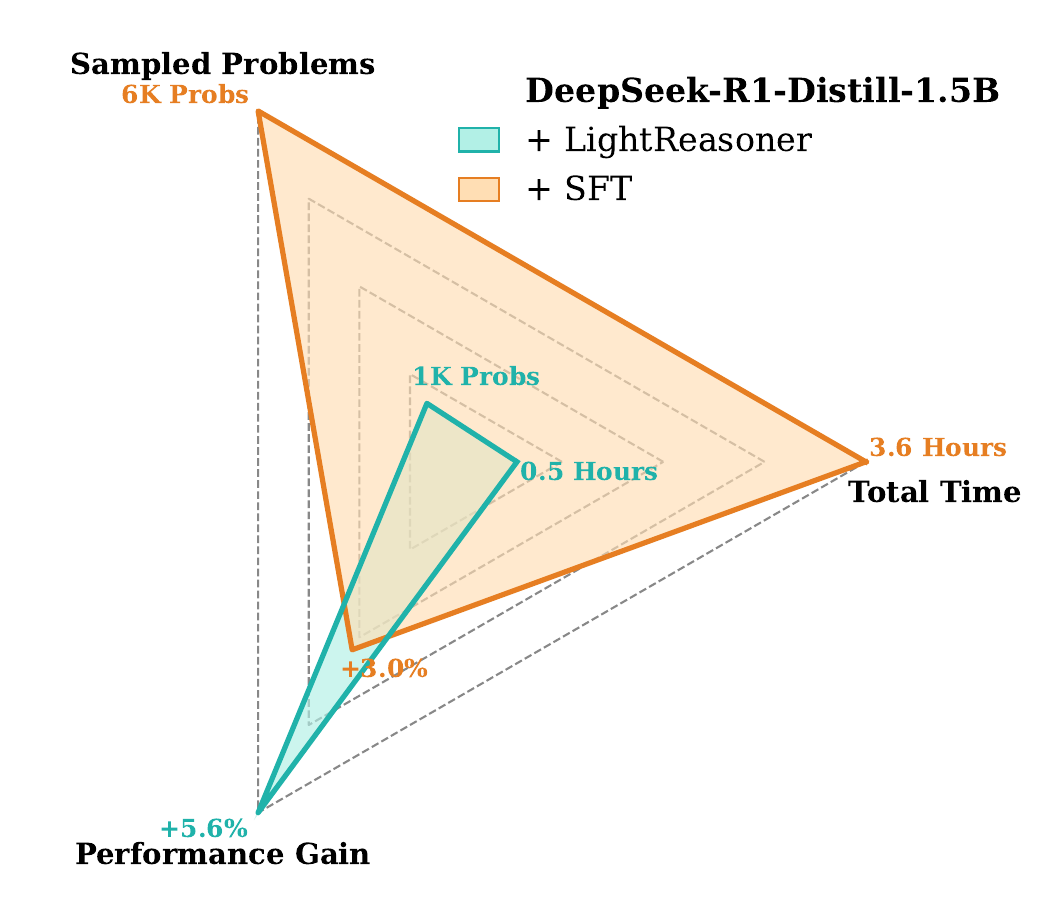}

        \scriptsize \textbf{(c)} DeepSeek-R1-1.5B
    \end{subfigure}\hfill    
    \begin{subfigure}[t]{0.5\linewidth}
        \centering
        \includegraphics[width=\linewidth]{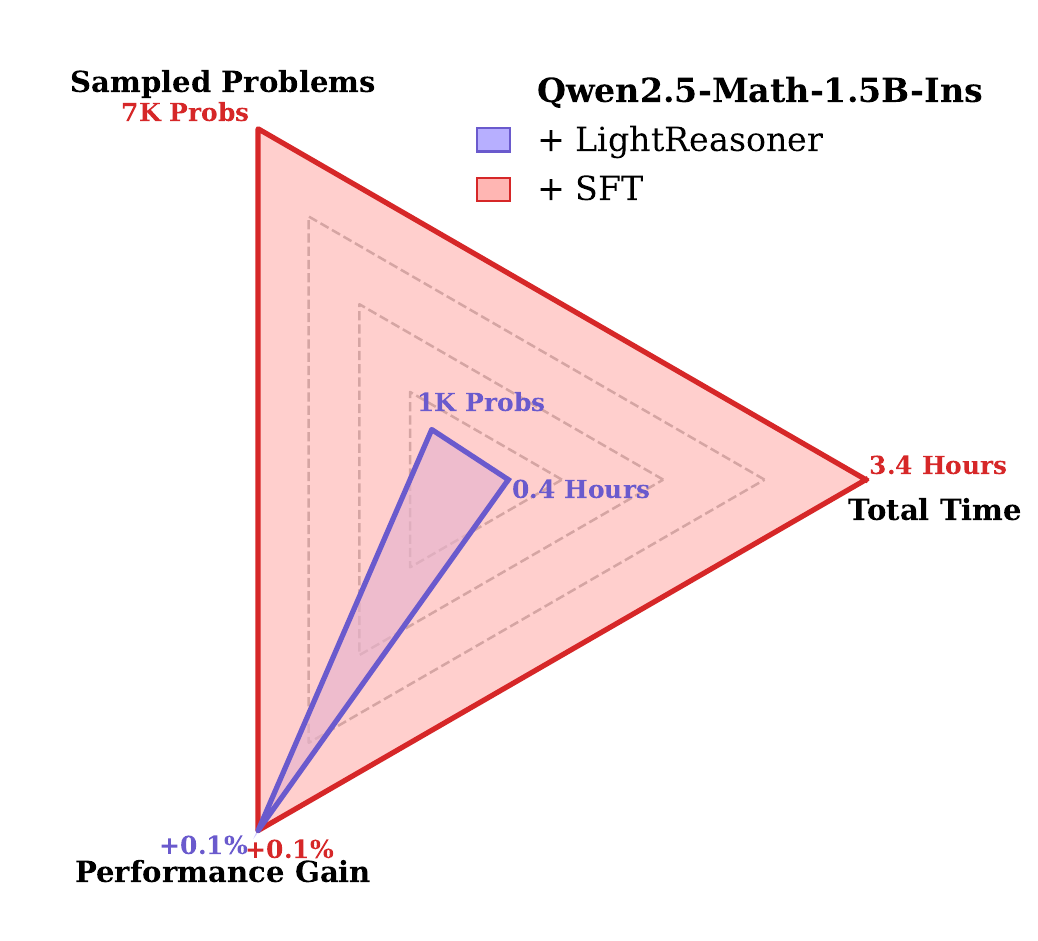}

        \scriptsize \textbf{(d)} Qwen2.5-Math-1.5B-Instruct
    \end{subfigure}

    \caption{\textbf{\model} achieves competitive or superior performance compared to \textbf{SFT} while substantially reducing resource consumption.}

    \label{fig:radar_comparison}
    
\end{figure}

These limitations have motivated more targeted strategies for improving LLM reasoning. Early as the Chain-of-Thought paper \citep{wei2022chain}, it was shown that LLMs acquire latent reasoning abilities during pre-training, which can be elicited through appropriate prompting. Recent works \citep{zhao2025learning, prabhudesai2025maximizing} show that models can refine themselves using internal feedback signals, such as self-certainty, without relying on external rewards or labeled data. Another line of research \citep{lin2024rho} finds that during pre-training, learning progress is concentrated on a small subset of tokens, suggesting that selective optimization on these high-impact tokens can yield significant gains in efficiency and performance. Together, these insights highlight the promise of harnessing underexplored internal learning dynamics to improve LLM reasoning. For completeness, a detailed discussion of related work is in \hyperref[app:related]{Appendix~\ref*{app:related}}.

Motivated by these insights, we pose a \textit{counterintuitive} possibility: can smaller, weaker models help guide the training of their larger, stronger counterparts? Specifically, we ask whether the divergent behaviors between a strong \textit{expert} model and a weaker \textit{amateur} model can pinpoint high-impact reasoning moments, and whether these moments can be transformed into effective training signals. In \hyperref[sec:tokeninfo]{\S\ref*{sec:tokeninfo}}, we analyze this Expert-Amateur dynamics through the Kullback–Leibler (KL) divergence between their next-token predictions: when models of different capabilities disagree strongly, those moments often mark critical steps in problem-solving. Our analysis shows that such steps, where the Expert confidently predicts the correct token while the Amateur diverges or shows uncertainty, are signaled by pronounced KL values. Exploiting these moments yields targeted supervision signals that reinforce the Expert’s reasoning strengths.

Building on these analyses, we introduce \textbf{\model}, a novel framework that leverages expert-amateur divergence to enhance LLM reasoning. \model\ operates in two stages: \textbf{(1) Sampling stage.} For each reasoning trajectory, Expert and Amateur models generate next-token predictions under identical prefixes. Steps where the KL divergence between their distributions exceeds a threshold are retained (\hyperref[sec:step_selection]{\S\ref*{sec:step_selection}}). From these selected steps, we construct supervision examples that encode Expert's \textit{relative} advantage by contrasting the two models' next-token distributions (\hyperref[sec:contrastive]{\S\ref*{sec:contrastive}}). \textbf{(2) Fine-tuning stage.} The \textit{same} Expert model is then trained to align with these contrastive signals, increasing probability of tokens where its advantage over Amateur is most pronounced, thereby reinforcing its strengths and avoiding amateur-like tendencies (\hyperref[sec:training]{\S\ref*{sec:training}}).

Our experimental evaluation highlights four key advantages of \model: 
$\bullet$ \textbf{Strong Performance Gains (\hyperref[sec:main_results]{\S\ref*{sec:main_results}}).} \model\ matches or surpasses SFT under equal experimental settings across 5 models and 7 benchmarks.
$\bullet$ \textbf{Order-of-Magnitude Efficiency (\hyperref[sec:efficiency]{\S\ref*{sec:efficiency}}).} \model\ delivers substantial savings: reducing total time cost by 90\%, requiring 80\% fewer sampled problems, and tuning 99\% fewer tokens compared to SFT\footnote{Unless otherwise stated, SFT in this paper refers to supervised fine-tuning on rejection-sampled model trajectories (also known as rejection sampling fine-tuning, RFT).} with rejection sampling, while entirely eliminating the need for ground-truth labels. $\bullet$ \textbf{Domain Expertise Drives Effective Contrasts (\hyperref[sec:pairing_effects]{\S\ref*{sec:pairing_effects}}).} Our analysis shows that the most effective Expert–Amateur contrasts stem from domain-specific expertise differences, reinforcing our insight that weaker models can provide meaningful contrast signals for teaching stronger ones.
$\bullet$ \textbf{Synergistic Design (\hyperref[sec:ablation]{\S\ref*{sec:ablation}}).} Ablation studies confirm the critical roles of step selection and contrastive supervision, along with their mutually-reinforcing effect, in enabling \model\ to strengthen the expert model’s reasoning efficiently and reliably.

\section{Methodology}
\label{sec:methodology}

\subsection{Preliminaries}

\paragraph{Autoregressive Language Model Generation.} 
Given vocabulary $\mathcal{A}$ and input $a_0$, a language model generates a response $a_{1:T} = [a_1, \ldots, a_T]$ autoregressively. At each step $t$, the model predicts the next token based on the prefix $s_t = [a_0, \ldots, a_{t-1}]$ and outputs distribution $\pi_{\mathrm{LM}}(\cdot \mid s_t)$ over $\mathcal{A}$. The joint likelihood factorizes as:
\begin{align}
    P(a_{1:T} \mid a_0) = \prod_{t=1}^{T} \pi_{\mathrm{LM}}(a_t \mid s_t).
\end{align}
The quality of reasoning emerges from the accumulation of individual token-level decisions. Thus, improving a model’s reasoning ability fundamentally amounts to refining its policy $\pi_{\mathrm{LM}}$. A central challenge, then, is to determine \emph{which} token-level decisions truly matter, and how to provide learning signals that target them effectively.

\paragraph{Learning from Behavioral Divergence.} 
Existing approaches often depend on human annotations or external verification mechanisms, which suffer from resource limitations, hindering their practicality for continuous improvement. We observe that models of different capabilities exhibit systematic differences in their token-level decision patterns. This insight motivates our approach: leveraging \emph{behavioral divergence} between models to automatically identify critical decision points and extract effective learning signals.

\paragraph{Critical Decision Points in Reasoning.}
The foundation of our approach lies in the observation that reasoning ability is shaped \textit{not} by uniform performance across all tokens, but by a handful of critical decision points. Cognitive science shows that certain reasoning steps exert disproportionate influence on final outcomes, creating natural bottlenecks in reasoning chains~\cite{chi1981categorization, ericsson1991toward}. Recent studies confirm that language models benefit most from training on such critical token subsets~\citep{lin2024rho, lin2024critical}. 

We therefore propose that targeting these high-stakes decision points creates a leverage effect: small improvements at bottlenecks can yield large overall gains. Our method operationalizes this intuition by exploiting expert–amateur differences to identify reasoning bottlenecks and focus enhancement where it matters most. To this end, we leverage two models with distinct reasoning capabilities: (1) an \textbf{Expert} model $\pi_E$, which we aim to improve, and (2) an \textbf{Amateur} model $\pi_A$, serving as a weaker baseline. As $\pi_E$ generates a response $a_{1:T}$ through a sequence of prefixes $s_{1:T}$, we evaluate both models at each step $t$ on the same prefix $s_t$. This produces paired distributions $(\pi_E(\cdot \mid s_t), \pi_A(\cdot \mid s_t))$, whose divergences reveal exactly where the Expert departs from amateur-level reasoning, providing targeted supervisory signals that concentrate on consequential decisions.

\begin{figure}[t]
    \centering
    \includegraphics[width=\linewidth, trim=0 7 0 0, clip]{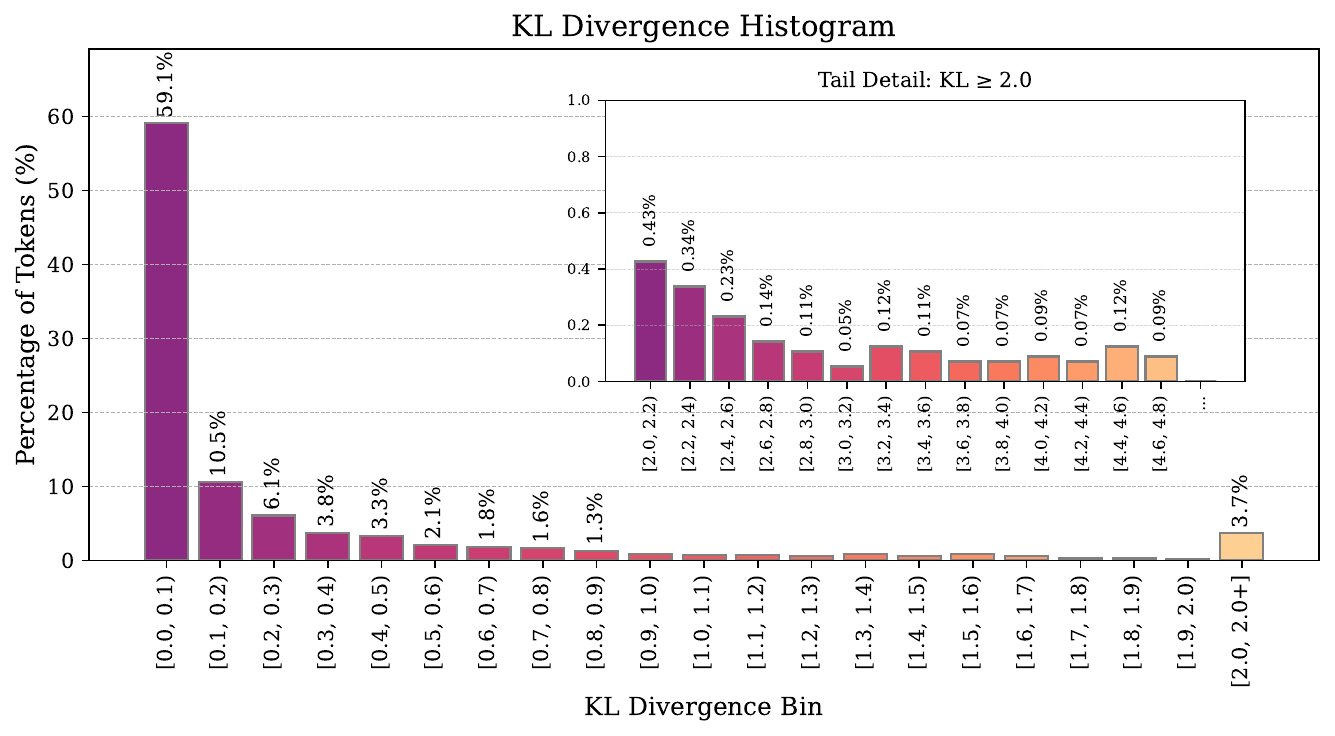}
    \caption{Most tokens show minimal KL divergence, \textbf{only few} exhibiting elevated values.}
    
     \label{fig:kl_histogram}
\end{figure}

\begin{figure}[t]
    \centering
    \includegraphics[width=\linewidth, trim=0 7 0 0, clip]{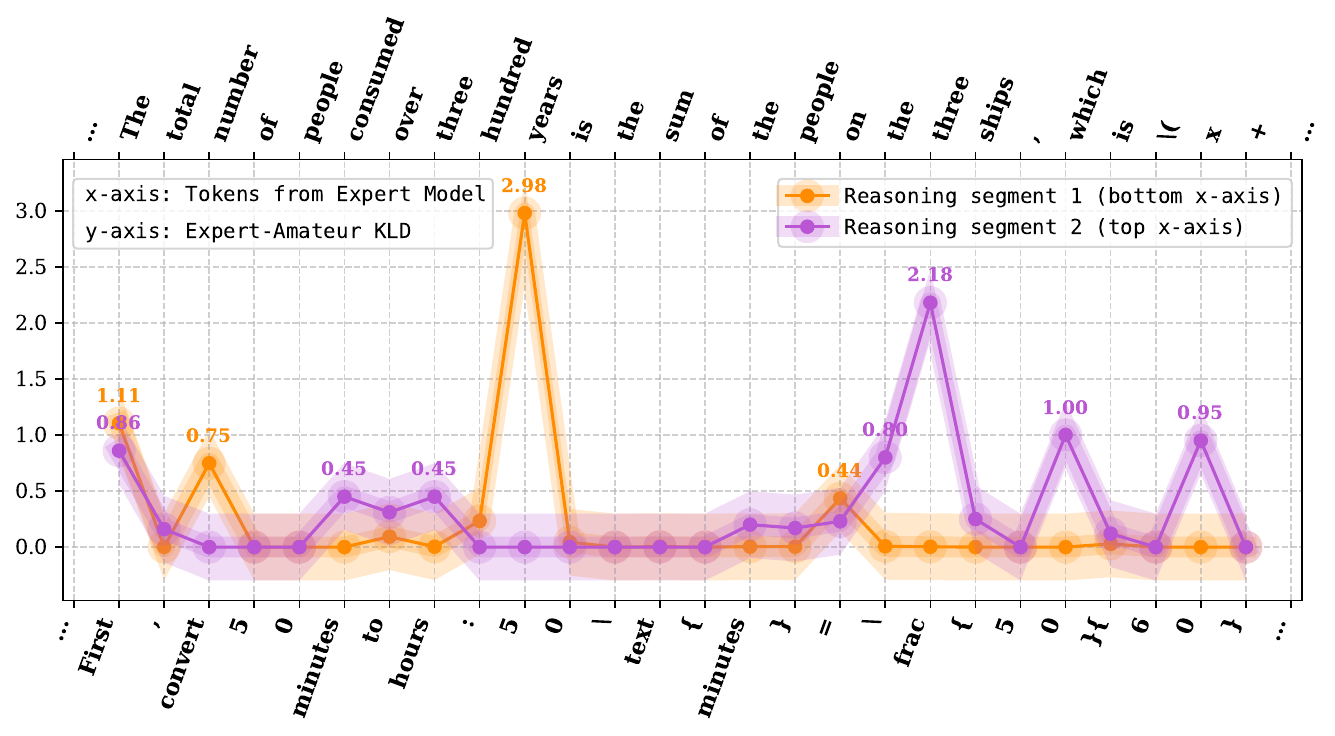}
    \caption{Predictable tokens yield near-zero KL divergence, while critical steps trigger notable spikes.}
    
    \label{fig:kl_trajectory_excerpt}
\end{figure}

\subsection{Token Informativeness}
\label{sec:tokeninfo}

To pinpoint critical decision points, we quantify the disagreement between Expert and Amateur models at each generation step $t$ using \textbf{Kullback--Leibler (KL) divergence}:
\begin{align}
&D_{\mathrm{KL}}\!\left(\pi_E(\cdot \mid s_t)\,\|\,\pi_A(\cdot \mid s_t)\right) \nonumber\\
&= \sum_{a \in \mathcal{A}} \pi_E(a \mid s_t)\,
   \log \frac{\pi_E(a \mid s_t)}{\pi_A(a \mid s_t)}.
\end{align}
Large KL values signals reasoning bottlenecks where the Expert departs sharply from amateur-level choices. In mathematical reasoning, these often recur at key transitional steps~\citep{ji2025first,prabhudesai2025maximizing}. Our analysis highlights three patterns that validate this perspective:

\begin{itemize}[leftmargin=*]

\item \textbf{Critical Tokens (\hyperref[fig:kl_histogram]{Figure~\ref*{fig:kl_histogram}}):} 
60\% of tokens exhibit negligible divergence $\mathrm{KLD} \in [0.0, 0.1)$, while only 20\% exceed 0.4, confirming that reasoning bottlenecks arise at specific moments.

\item \textbf{Reasoning Complexity (\hyperref[fig:kl_trajectory_excerpt]{Figure~\ref*{fig:kl_trajectory_excerpt}}):} 
Case studies show that KL divergence stays near zero for routine tokens but spikes at demanding steps such as arithmetic operations and logical transitions, tracking underlying complexity.

\item \textbf{Hidden Disagreements:} 
When the Expert and Amateur disagree on top-1 tokens, the average KL jumps to 1.99, versus 0.166 under agreement. Even among top-1 agreements, 10\% of cases still show $\mathrm{KLD} > 0.4$, exposing hidden misalignments beneath apparent consensus.

\end{itemize}

\begin{figure*}[t]
    \centering
    \includegraphics[width=\textwidth, trim=0 110 80 0, clip]{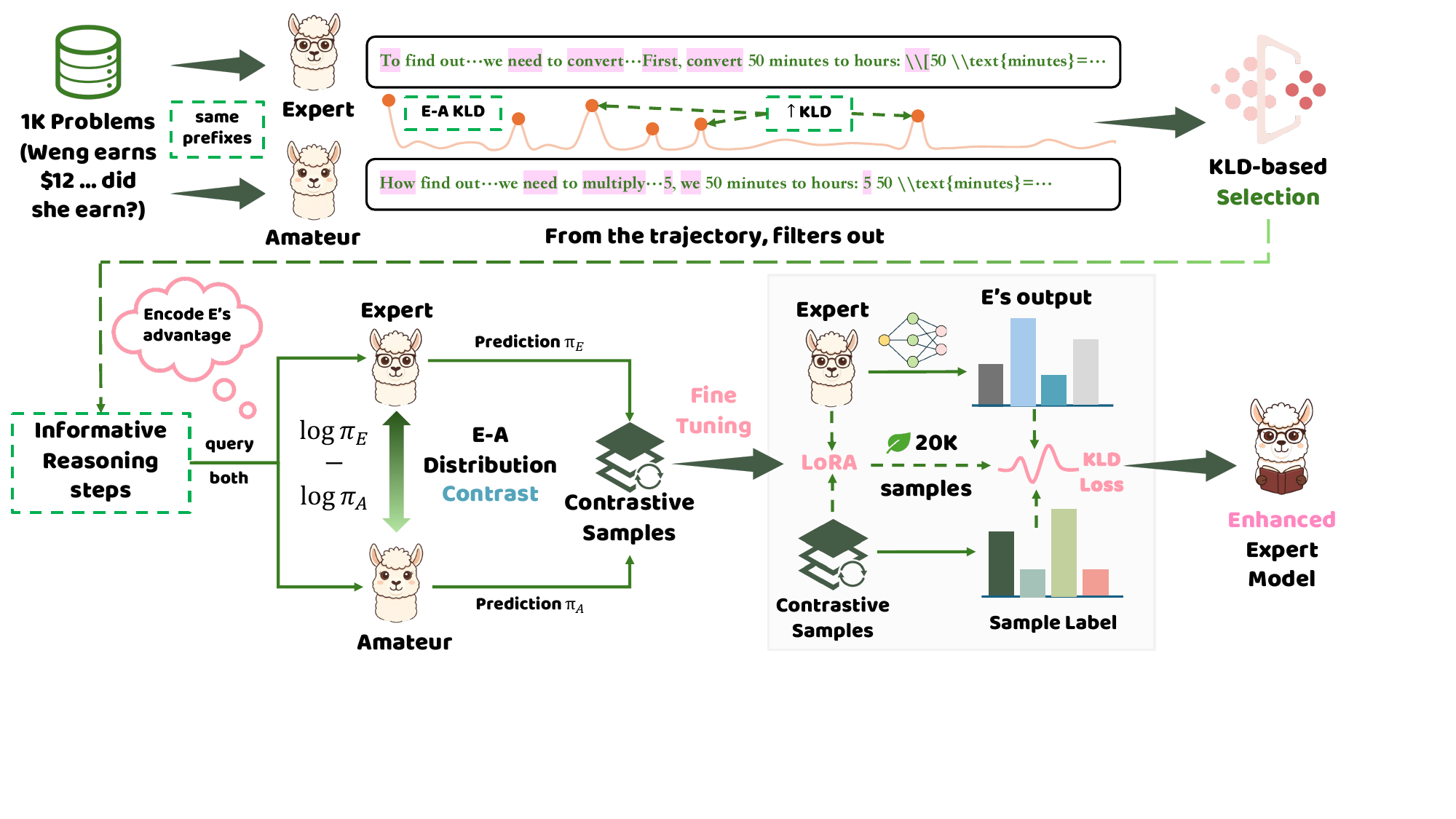}
    \caption{
        Overview of the \textbf{LightReasoner} framework.
        \textbf{Sampling Stage}: Expert and Amateur models generate distributions $\pi_E$ and $\pi_A$.
        Informative step selection (\hyperref[sec:step_selection]{\S\ref*{sec:step_selection}}) retains steps with $D_{\text{KL}}(\pi_E \parallel \pi_A) > \beta$, and contrastive supervision (\hyperref[sec:contrastive]{\S\ref*{sec:contrastive}}) constructs soft labels $v_C$ capturing the Expert's advantage through Expert–Amateur contrast.
        \textbf{Fine-tuning Stage}: The Expert model is enhanced by minimizing the KLD between its output and $v_C$ (\hyperref[sec:training]{\S\ref*{sec:training}}).
    }
    
    \label{fig:lr_frame}
\end{figure*}

\subsection{\model\ Framework}

We present \textbf{\model} (\hyperref[fig:lr_frame]{Figure~\ref*{fig:lr_frame}}), a novel self-distillation framework for enhancing LLM reasoning. It is built on \textit{three} core components: 
(1) an informative step-selection mechanism that pinpoints critical reasoning steps via divergence-based metrics (\hyperref[sec:step_selection]{\S\ref*{sec:step_selection}}); 
(2) a contrastive method for constructing supervision samples that capture the Expert model’s reasoning advantage (\hyperref[sec:contrastive]{\S\ref*{sec:contrastive}}); 
(3) a training objective that distills the learning signal back into the Expert (\hyperref[sec:training]{\S\ref*{sec:training}}). 
Full procedure in \hyperref[alg:lr]{Algorithm~\ref*{alg:lr}}.

\subsubsection{Informative Step Selection}
\label{sec:step_selection}

Reasoning trajectories consist of tokens with uneven learning value. Routine steps often yield close Expert–Amateur agreement, while valuable supervision arises at critical decision points where expert reasoning is decisive. 

To capture these moments, we introduce \emph{informative step selection}, which uses model KL divergence to quantify reasoning criticality: large divergences mark bottlenecks where expert knowledge can be deterministic. We implement this via \emph{$\beta$-filtering}, retaining only steps where the Expert’s advantage is pronounced. For a given prefix $s_t$, a step is selected if
\begin{align}
    D_{\mathrm{KL}}\!\left(\pi_E(\cdot \mid s_t) ~\|~ \pi_A(\cdot \mid s_t)\right) > \beta.
\end{align}

\subsubsection{Contrastive Distributional Supervision}
\label{sec:contrastive}

After identifying informative steps, we convert Expert–Amateur disagreement into effective training signals. Instead of one-hot targets, which discard distributional information, we construct \emph{soft} targets that encode the Expert’s \emph{relative} advantage. For a step $s_t$ passing the $\beta$-filter, we first define the \emph{masked support set} of model vocabulary $\mathcal{A}$:

\begin{equation}
\begin{split}
\mathcal{A}_{\text{mask}}
&= \Bigl\{
  a \in \mathcal{A}:\,
  \pi_E(a \mid s_t) \ge \\
  &\hspace{2.5em}
  \alpha \cdot 
  \max_{b \in \mathcal{A}}
  \pi_E(b \mid s_t)
  \Bigr\},
\end{split}
\end{equation}
where $\alpha \in (0, 1]$ \citep{li2022contrastive} removes low-confidence tokens from the vocabulary, preventing noisy tail probabilities from distorting the supervision signal. For each $a \in \mathcal{A}_{\text{mask}}$, we compute the unnormalized \emph{contrast score} \citep{li2022contrastive}:
\begin{equation}
    v'_C(a \mid s_t)
    = \log\!\frac{\pi_E(a \mid s_t)}{\pi_A(a \mid s_t)},
\end{equation}
quantifying the Expert's advantage margin. 

We apply \textit{softmax} to obtain a normalized distribution $\tilde{v}_C$, then extend it over all tokens $a \in \mathcal{A}$:
\begin{align}
    v_C(a \mid s_t) 
    &= \tilde{v}_C(a \mid s_t) \cdot \mathbbm{1}\!\left[ a \in \mathcal{A}_{\text{mask}} \right], 
\end{align}
The resulting $v_C(\cdot \mid s_t)$ encodes the expert superiority as probabilistic supervision, enabling targeted refinement that reinforces expert-like decisions.

\subsubsection{Self-Distillation Training Objective}
\label{sec:training}

The framework is completed with a self-distillation objective that transfers the signal encoded in $v_C$ back into the Expert model. Specifically, the Expert distribution $\pi_E$ is trained to match $v_C$ by minimizing the KL divergence:
\begin{equation}
\begin{split}
\mathcal{L}(s_t) 
&= D_{\mathrm{KL}}\!\big(v_C(\cdot \mid s_t) \,\big\|\, \pi_E(\cdot \mid s_t)\big) \\[3pt]
&= \sum_{a \in \mathcal{A}} 
   v_C(a \mid s_t) 
   \log\!\frac{v_C(a \mid s_t)}{\pi_E(a \mid s_t)}.
\end{split}
\label{eq:kl-full}
\end{equation}
Since the first term is constant with respect to $\pi_E$, \eqref{eq:kl-full} is equivalent to cross-entropy minimization:
\begin{align}
\tilde{\mathcal{L}}(s_t) 
&= -\sum_{a \in \mathcal{A}} v_C(a \mid s_t) \,\log \pi_E(a \mid s_t),
\end{align}
which reinforces the Expert's probability mass on tokens where its advantage over the Amateur is most pronounced. 

Detailed math derivations, analysis of component synergy, and an intuitive explanation of how the method enhances reasoning are provided in \hyperref[app:kl2contrast]{Appendix~\ref*{app:kl2contrast}} and \hyperref[app:connect3]{Appendix~\ref*{app:connect3}}.

\section{Experiment}
\label{sec:experiment}


\begin{table*}[t]
\centering
\renewcommand{\arraystretch}{1.0}
\setlength{\tabcolsep}{6pt}

\begin{tabular}{lcccccccc}
\toprule
\addlinespace[0.1em]
\vcenteredcell{\textbf{Model}} &
\vcenteredcell{\textbf{GSM8K}} &
\vcenteredcell{\textbf{MATH}} &
\vcenteredcell{\textbf{SVAMP}} &
\vcenteredcell{\textbf{ASDiv}} &
\vcenteredcell{\textbf{\shortstack{Minerva\\Math}}} &
\vcenteredcell{\textbf{\shortstack{Olympiad\\Bench}}} &
\vcenteredcell{\textbf{\shortstack{MMLU\\STEM}}} &
\vcenteredcell{\textbf{AVG.}} \\
\addlinespace[0.1em]
\midrule
\addlinespace[0.1em]
\multicolumn{9}{l}{\textbf{\textit{Qwen2.5-Math-1.5B}}} \\
\quad Baseline & 42.5 & 34.2 & 68.8 & 68.1 & 9.9 & 23.7 & 49.8 & 42.4\\
\hdashline
\quad \cellcolor{softPink}\textbf{+ SFT} & 69.2 & 57.1 & 64.1 & 70.2 & \textbf{15.1} & \textbf{27.6} & 47.7 & 50.1\\
\quad \cellcolor{softGreen}\textbf{+ LightR} & \textbf{70.6} & \textbf{59.3} & \textbf{76.0} & \textbf{79.8} & 11.4 & 27.1 & \textbf{54.9} & \textbf{54.2}\\
\midrule
\addlinespace[0.1em]
\multicolumn{9}{l}{\textbf{\textit{Qwen2.5-Math-1.5B-Instruct}}} \\
\quad Baseline & 84.8 & 75.8 & 94.2 & 94.7 & 29.4 & 37.5 & 57.4 & 67.7\\
\hdashline
\quad \cellcolor{softPink}\textbf{+ SFT} & 85.4 & 75.8 & 93.5 & 94.7 & 31.6 & 37.5 & 56.2 & 67.8\\
\quad \cellcolor{softGreen}\textbf{+ LightR} & \textbf{86.7} & 75.5 & 93.0 & 94.1 & \textbf{32.0} & \textbf{37.8} & 55.2 & \textbf{67.8}\\
\midrule
\addlinespace[0.1em]
\multicolumn{9}{l}{\textbf{\textit{DeepSeek-R1-Distill-Qwen-1.5B}}} \\
\quad Baseline & 75.2 & 54.2 & 79.9 & 84.9 & 16.2 & 19.1 & 22.3 & 50.3\\
\hdashline
\quad \cellcolor{softPink}\textbf{+ SFT} & 78.2 & \textbf{60.3} & 81.5 & 87.4 & \textbf{18.4} & 21.2 & 26.2 & 53.3\\
\quad \cellcolor{softGreen}\textbf{+ LightR} & \textbf{79.5} & 60.2 & \textbf{83.5} & \textbf{87.5} & 18.0 & \textbf{36.5} & \textbf{26.2} & \textbf{55.9}\\
\midrule
\addlinespace[0.1em]
\multicolumn{9}{l}{\textbf{\textit{Qwen2.5-Math-7B}}} \\
\quad Baseline & 57.5 & 51.8 & 67.9 & 72.7 & 14.0 & 16.0 & 69.8 & 50.0\\
\hdashline
\quad \cellcolor{softPink}\textbf{+ SFT} & 64.4 & \textbf{63.3} & 76.2 & 76.6 & 12.1 & \textbf{20.5} & 68.5 & 54.5\\
\quad \cellcolor{softGreen}\textbf{+ LightR} & \textbf{67.9} & 57.8 & \textbf{77.2} & \textbf{80.6} & 12.1 & 16.9 & \textbf{70.5} & \textbf{54.7}\\
\midrule
\addlinespace[0.1em]
\multicolumn{9}{l}{\textbf{\textit{Qwen2.5-Math-7B-Instruct}}} \\
\quad Baseline & 95.2 & 83.2 & 93.9 & 95.3 & 33.8 & 41.5 & 69.3 & 73.2\\
\hdashline
\quad \cellcolor{softPink}\textbf{+ SFT} & 95.4 & 83.1 & \textbf{94.1} & 95.2 & \textbf{38.2} & 40.7 & 68.2 & \textbf{73.6}\\
\quad \cellcolor{softGreen}\textbf{+ LightR} & \textbf{95.8} & \textbf{83.6} & 93.1 & 95.2 & 34.2 & 39.0 & 67.8 & 72.7\\
\bottomrule
\end{tabular}

\caption{\textbf{Zero-shot pass@1}\zsref\ accuracy (\%) across evaluation benchmarks. \model\ consistently achieves comparable or superior performance over SFT across 5 baseline models and 7 mathematical benchmarks.}

\label{tab:main_results}
\end{table*}


\begin{table}[t]
\centering
\footnotesize
\setlength{\tabcolsep}{6.5pt}
\renewcommand{\arraystretch}{1.05}

\vspace{-0.1in}

\begin{tabular}{lccccc}
\toprule
\textbf{Method} & \textbf{Time} & \textbf{Probs.} & \textbf{Tokens} & \textbf{Gain} \\
\midrule
\multicolumn{5}{l}{\textit{Qwen2.5-Math-1.5B}} \\
\quad + SFT & 4.0h & 3952 & 1.77M & +7.7\% \\
\rowcolor{softGreen}\quad \textbf{+ LightR} & \textbf{0.5h} & \textbf{1000} & \textbf{0.02M} & \textbf{+11.8\%} \\
\midrule
\multicolumn{5}{l}{\textit{Qwen2.5-Math-7B}} \\
\quad + SFT & 9.5h & 6029 & 2.20M & +4.5\% \\
\rowcolor{softGreen}\quad \textbf{+ LightR} & \textbf{0.75h} & \textbf{1000} & \textbf{0.02M} & \textbf{+4.7\%} \\
\midrule
\multicolumn{5}{l}{\textit{DeepSeek-R1-Distill-Qwen-1.5B}} \\
\quad + SFT & 3.6h & 6023 & 5.95M & +3.0\% \\
\rowcolor{softGreen}\quad \textbf{+ LightR} & \textbf{0.5h} & \textbf{1000} & \textbf{0.02M} & \textbf{+5.6\%} \\
\midrule
\multicolumn{5}{l}{\textit{Qwen2.5-Math-1.5B-Instruct}} \\
\quad + SFT & 3.4h & 7153 & 2.08M & +0.1\% \\
\rowcolor{softGreen}\quad \textbf{+ LightR} & \textbf{0.4h} & \textbf{1000} & \textbf{0.02M} & \textbf{+0.1\%} \\
\bottomrule
\end{tabular}

\caption{\textbf{Efficiency comparison} across time, sampled problems, tuned tokens, and average performance gain.}
\vspace{-0.1in}

\label{tab:efficiency_comparison}
\end{table}


\begin{table}[t]
\centering
\footnotesize
\setlength{\tabcolsep}{6pt}
\renewcommand{\arraystretch}{1.05}

\vspace{-0.1in}

\begin{tabular*}{\linewidth}{@{\extracolsep{\fill}} l c c c @{}}
\toprule
\textbf{Attribute} & \textbf{Efficiency} & \textbf{SFT} & \textbf{LightR} \\
\midrule
Full sampling  & \downarroweff & \cmark & \xmark \\
All-token tuning   & \downarroweff & \cmark & \xmark \\
GT verification    & \downarroweff & \cmark & \xmark \\
\hdashline
Prefix sampling & \uparroweff   & \xmark & \cmark \\
Selective tuning   & \uparroweff   & \xmark & \cmark \\
Verification-free  & \uparroweff   & \xmark & \cmark \\
\bottomrule
\end{tabular*}

\caption{\textbf{SFT vs. \model} – efficiency comparison at a glance. \uparroweff\ and \downarroweff\ indicate whether each aspect improves or reduces efficiency.}


\label{tab:sft_vs_lightreasoner_inline}
\end{table}

\begin{figure}[t]
    \centering
    \begin{subfigure}[t]{\linewidth}
        \centering
        \includegraphics[width=\linewidth]{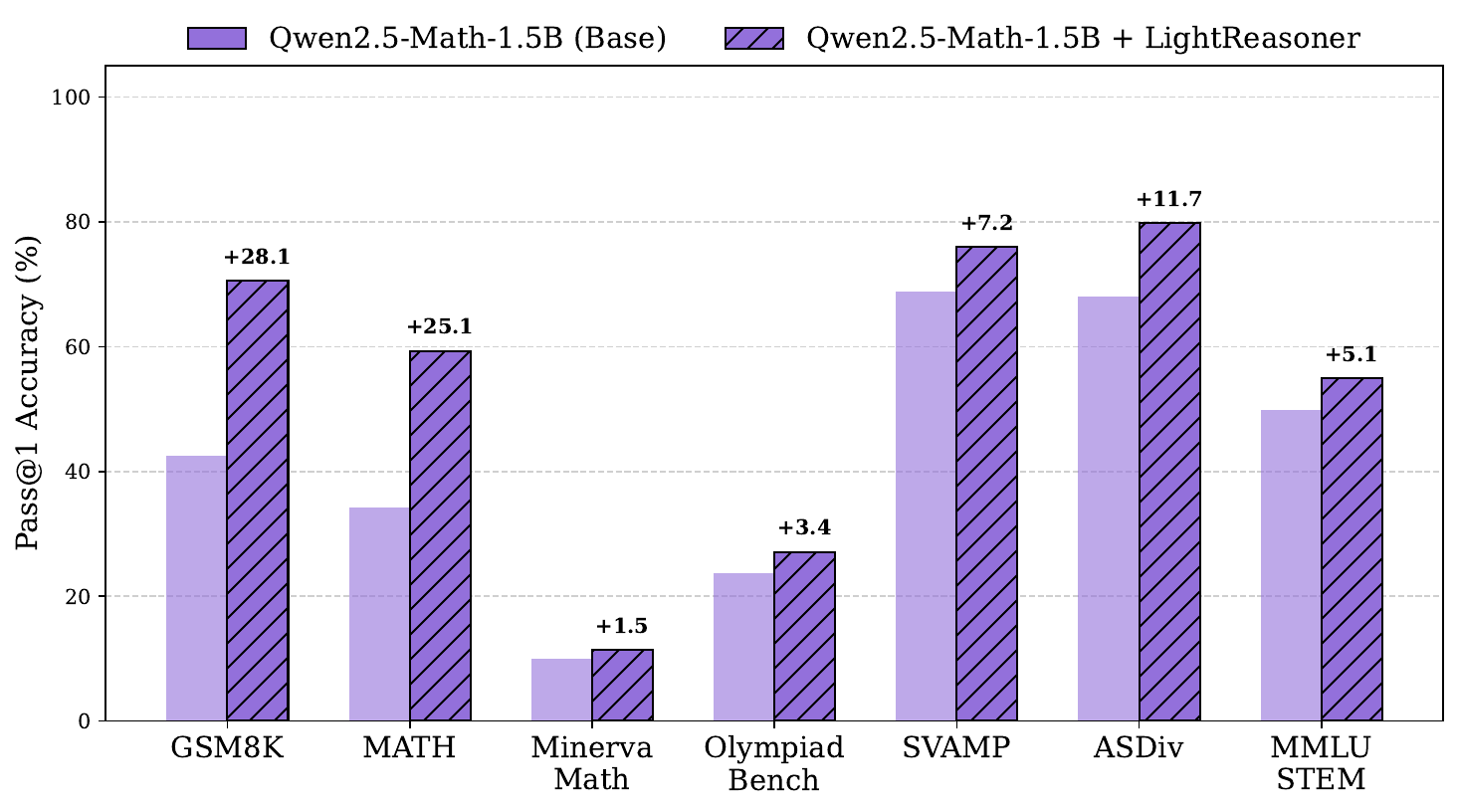}
    \end{subfigure}
    
    \begin{subfigure}[t]{\linewidth}
        \centering
        \includegraphics[width=\linewidth]{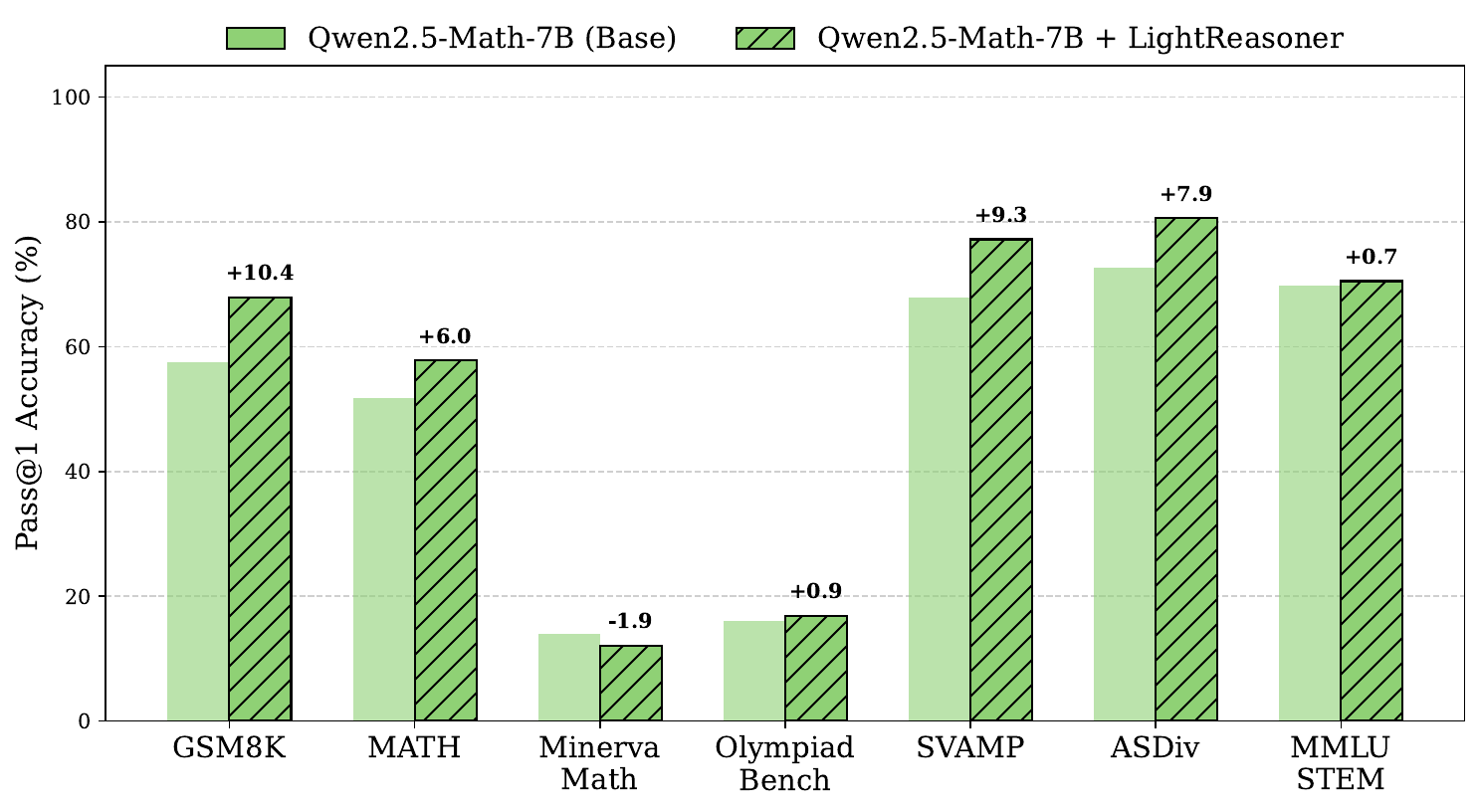}
    \end{subfigure}

    \caption{\textbf{\model} consistently improves accuracy across 7 math benchmarks for baseline models.}
    \label{fig:lightr_bars_combined}
\end{figure}

To assess the effectiveness of \model, we structure our experiments around four research questions:  
$\bullet$ \textbf{RQ1:} How does \model\ improve performance across different baseline models?  
$\bullet$ \textbf{RQ2:} How much does \model\ reduce computational costs?  
$\bullet$ \textbf{RQ3:} Which factors drive successful Expert–Amateur collaboration?  
$\bullet$ \textbf{RQ4:} How do the core components of \model\ contribute to its effectiveness?

\subsection{Experimental Setup}

\subsubsection{Models and Data}
\label{sec:model_data}
\model\ leverages Expert–Amateur model pairing to generate supervision signals. The Expert models include Qwen2.5-Math-1.5B and 7B, their Instruct versions, and the DeepSeek-R1-Distill variant~\citep{yang2024qwen2, guo2025deepseek}. The Amateur is fixed as Qwen2.5-0.5B~\citep{yang2024qwen2}, a base model with general linguistic reasoning ability but without specialized math training.

For generating supervision samples, we use the GSM8K training set~\citep{cobbe2021training}, selected for its emphasis on step-by-step, broadly applicable logical reasoning. CoT prompting~\citep{wei2022chain} is employed to elicit reasoning trajectories. Comprehensive model specifications and dataset descriptions are provided in \hyperref[app:datasets]{Appendix~\ref*{app:datasets}} and \hyperref[app:models]{Appendix~\ref*{app:models}}.

\subsubsection{Training Configuration}
\label{sec:training_config}
We fine-tune the \textit{same} Expert model on the curated \model\ supervision set using LoRA~\citep{hu2022lora} for parameter-efficient training. Key hyperparameters are as follows: a masking threshold of $\alpha = 0.2$ to filter out low-probability Expert tokens~\citep{li2022contrastive}; a KL divergence threshold of $\beta = 0.4$ to isolate informative reasoning steps; and the sampling rollout length capped at 128 tokens (ablation on the sampling length is provided in \hyperref[app:truncation]{Appendix~\ref*{app:truncation}}), as early steps are typically less error-prone~\citep{ji2025first}. Models are fine-tuned for 1000 steps, with each step comprising 16 contrastive supervision samples, which we find sufficient for convergence. Full configuration details are provided in \hyperref[app:lr_extra]{Appendix~\ref*{app:lr_extra}}.

\subsubsection{Design Rationale}
\label{sec:design_rationale}

Despite the effectiveness of the step selection mechanism (\hyperref[sec:step_selection]{\S\ref*{sec:step_selection}}), one potential concern arises when both models follow an incorrect reasoning path. In such cases, the resulting KL divergence may still be large, allowing these steps to pass the filtering and enter the supervision set as false positives. This reflects shared errors rather than genuine expertise differences, potentially reinforcing erroneous patterns in the Expert model. 

We mitigate this risk through two design choices. First, we use GSM8K to generate supervision samples, as its step-wise reasoning and basic arithmetic largely fall within the Expert model’s capability, reducing the likelihood of systematic failure. At the same time, it ensures that the Amateur, despite lacking domain-specific training, can still produce coherent outputs, enabling meaningful contrast with the Expert. Second, we restrict sampling to short rollout prefixes rather than full trajectories, as prior work \citep{ji2025first} shows that early reasoning steps are significantly more stable, while later steps are more prone to cascading errors.

\subsubsection{Evaluation}
\label{sec:eval}
For competition, we implement SFT on the baseline models (\hyperref[app:sft_extra]{Appendix~\ref*{app:sft_extra}}), LoRA fine-tuning on demonstrations of correct reasoning trajectories collected via rejection sampling~\citep{yuan2023scalingrelationshiplearningmathematical}. We evaluate the baseline models, the SFT-trained variants, and the \model\ fine-tuned counterparts on a diverse suite of benchmarks ranging from basic arithmetic to expert-level math reasoning (\hyperref[app:datasets]{Appendix~\ref*{app:datasets}}). All evaluations are conducted with the Qwen2.5-Math toolkit\footnote{\url{https://github.com/QwenLM/Qwen2.5-Math}}, and we report \textbf{zero-shot pass@1}\zsref as the primary metric.

\begin{figure}[t]
    \centering
    \includegraphics[width=\linewidth]{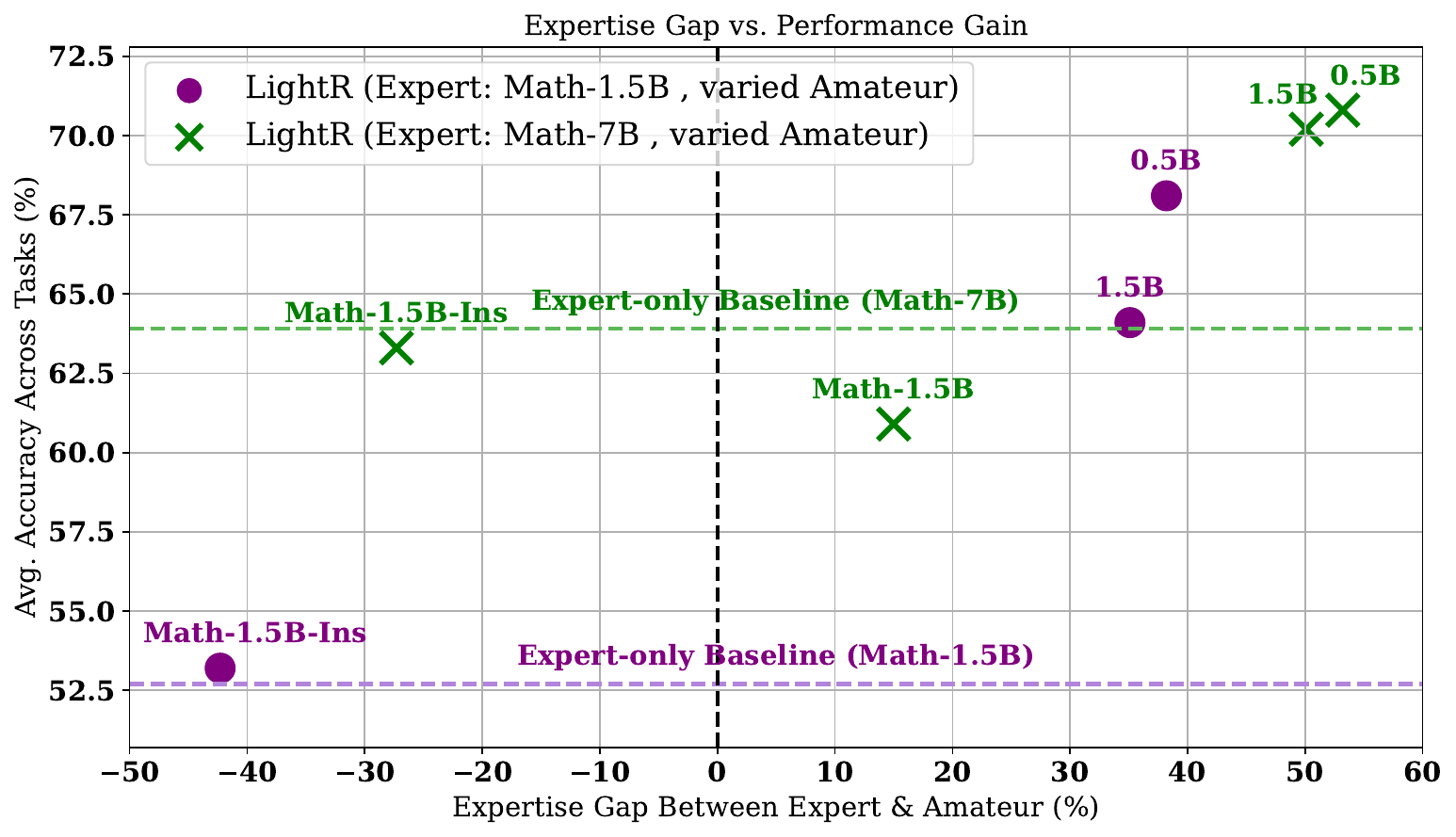}

    \caption{Performance gains from \model\ \textbf{diminish} as the Expert-Amateur expertise \textbf{gap narrows}. Label above each data point denotes the Amateur model paired with the Expert.}
    
    \label{fig:gap_vs_perf}
\end{figure}


\begin{table}[t]
\centering
\footnotesize
\setlength{\tabcolsep}{6pt}
\renewcommand{\arraystretch}{1.05}

\vspace{0.3em}
\begin{tabular*}{\linewidth}{@{\extracolsep{\fill}} l c c c @{}}
\toprule
\textbf{Attribute} & \textbf{Utility} & \textbf{CD} & \textbf{LightR} \\
\midrule
Contrast usage        & /           & Inference  & Training \\
\hdashline
Size-based contrast   & \downarroweff & \cmark & \xmark \\
Expertise contrast    & \uparroweff   & \xmark & \cmark \\
\hdashline
Persistent improvement   & \uparroweff   & \xmark & \cmark \\
Independent inference & \uparroweff   & \xmark & \cmark \\
\bottomrule
\end{tabular*}

\caption{\textbf{Key differences} between Contrastive Decoding (CD) and \model. \uparroweff\ and \downarroweff\ indicate whether each attribute improves or reduces the method's practical applicability.}

\label{tab:cd_vs_lightreasoner}
\end{table}

\subsection{Performance Improvements (RQ1)}
\label{sec:main_results}

\hyperref[tab:main_results]{Table~\ref*{tab:main_results}} presents the \textbf{zero-shot pass@1}\zsfootnote\ accuracy across a diverse suite of math reasoning benchmarks. Across 5 models and 7 datasets, \model\ delivers consistent performance improvements and demonstrates fundamental advances in math reasoning capability. These results highlight \model's strong generalization, which we analyze through three perspectives below.

\begin{itemize}[leftmargin=*]

\item \textbf{Cross-Dataset Reasoning Enhancement.} Despite being trained exclusively on GSM8K, \model\ achieves consistent gains across diverse benchmarks including MATH, SVAMP, and ASDiv. This shows that \model\ cultivates foundational reasoning enhancement rather than memorizing dataset-specific heuristics. By focusing on reasoning bottlenecks where the expert diverges from amateur patterns, the method captures transferable logical structures that extend beyond the training domain.

\item \textbf{Adaptive Enhancement across Models.} Our approach delivers consistent improvements across models of different capacities. For non-instruct models (e.g., Qwen2.5-Math-1.5B), we observe dramatic gains (+28.1\% on GSM8K; +25.1\% on MATH), showing that contrastive supervision can activate latent reasoning circuits previously dormant. For heavily optimized instruct models (e.g., 1.5B-Instruct), the improvements are modest but steady (+1.9\% on GSM8K; +2.6\% on Minerva Math), suggesting our method refines existing reasoning pathways. This differential effectiveness highlights how the method adapts flexibly to varying model capabilities.

\item \textbf{Superior Efficiency over SFT.} Direct comparisons show that \model\ matches or surpasses SFT performance while consuming 90\% less total time, 80\% fewer sampled problems, and 99\% fewer tuned tokens (\hyperref[sec:efficiency]{\S\ref*{sec:efficiency}}). These efficiency gains stem from two synergistic components: step selection (\hyperref[sec:step_selection]{\S\ref*{sec:step_selection}}), which concentrates learning on the $\sim$20\% of tokens that drive reasoning (\hyperref[sec:tokeninfo]{\S\ref*{sec:tokeninfo}}), and contrastive supervision (\hyperref[sec:contrastive]{\S\ref*{sec:contrastive}}), which constructs training signals that encode the expert’s advantage over amateurish tendencies.

\end{itemize}


\begin{table*}[t]
\centering
\renewcommand{\arraystretch}{1.0}
\setlength{\tabcolsep}{6pt}

\begin{tabular}{lccccccc}
\toprule
\addlinespace[0.1em]
\vcenteredcell{\textbf{Amateur Model}} &
\vcenteredcell{\textbf{$\Delta$ Perf.}} &
\vcenteredcell{\textbf{GSM8K}} &
\vcenteredcell{\textbf{MATH}} &
\vcenteredcell{\textbf{SVAMP}} &
\vcenteredcell{\textbf{ASDiv}} &
\vcenteredcell{\textbf{\shortstack{MMLU\\STEM}}} &
\vcenteredcell{\textbf{AVG.}} \\
\addlinespace[0.1em]
\midrule

\multicolumn{7}{l}{\textbf{\textit{Expert: Qwen2.5-Math-1.5B}}} \\
\rowcolor{lemon} \quad Qwen2.5-0.5B                & 38.2  & 70.6 & 59.3 & 76.0 & 79.8 & 54.9 & \textbf{68.1} \\
\quad Qwen2.5-1.5B                 & 35.1 & 63.4 & 57.1 & 69.7 & 75.7 & 54.8 & 64.1 \\
\quad Qwen2.5-Math-1.5B-Ins        & -42.3 & 41.4 & 35.5 & 67.5 & 66.4 & 55.0 & 53.2 \\
\hdashline
\quad \textit{Expert Only (Baseline)}             & / & 42.5 & 34.2 & 68.8 & 68.1 & 49.8 & 52.7 \\

\midrule

\multicolumn{7}{l}{\textbf{\textit{Expert: Qwen2.5-Math-7B}}} \\
\rowcolor{lemon} \quad Qwen2.5-0.5B                 & 53.2 & 67.9 & 57.8 & 77.2 & 80.6 & 70.5 & \textbf{70.8} \\
\quad Qwen2.5-1.5B                 & 50.1 & 69.0 & 56.0 & 77.6 & 78.9 & 69.5 & 70.2 \\
\quad Qwen2.5-Math-1.5B            & 15.0 & 56.9 & 50.2 & 63.5 & 63.4 & 70.7 & 60.9 \\
\quad Qwen2.5-Math-1.5B-Ins        & -27.3 & 59.4 & 49.0 & 68.3 & 69.6 & 70.3 & 63.3 \\
\hdashline
\quad \textit{Expert Only (Baseline)}              & / & 57.5 & 51.8 & 67.9 & 72.7 & 69.8 & 63.9 \\
\bottomrule
\end{tabular}

\caption{\textbf{Impact of expertise-driven contrast.} 
We fix the Expert model and vary the Amateur; \(\Delta \text{Perf.}\) denotes the Expert-Amateur performance difference on GSM8K; each group ends with the Expert baseline.}

\label{tab:amateur_variation}
\end{table*}

\subsection{Efficiency Study (RQ2)}
\label{sec:efficiency}

We assess efficiency along three key dimensions:
(1) \textbf{Time budget}: the total sampling and fine-tuning time on a single NVIDIA H200 GPU without inference accelerators;  
(2) \textbf{Training instances}: the number of distinct training set problems to generate the supervision dataset;  
(3) \textbf{Tuned tokens}: the computational overhead at the token level, where \model\ trains on selective next-token predictions while SFT optimizes over full reasoning trajectories.  
As shown in \hyperref[tab:efficiency_comparison]{Table~\ref*{tab:efficiency_comparison}} and \hyperref[fig:radar_comparison]{Figure~\ref*{fig:radar_comparison}}, \model\ consistently outperforms SFT with only a fraction of the resources. These efficiency improvements stem from three core design principles of \model\ that directly address SFT’s computational bottlenecks:

\begin{itemize}[leftmargin=*]

\item \textbf{Sampling efficiency via prefix termination.} 
During rejection sampling, SFT must generate complete reasoning trajectories. \model, on the other hand, halts each sampling rollout at 128 tokens, sharply reducing the cost. This design leverages the observation that early reasoning steps provide more reliable signals with fewer cascading errors~\citep{ji2025first}, avoiding computational expense on error-prone later steps.

\item \textbf{Training efficiency through selective tokens.} 
\model\ concentrates learning on high-value reasoning moments rather than low-return tokens. This selectivity accounts for the large gap in token usage between SFT and \model. By avoiding SFT’s indiscriminate tuning over full trajectories, \model\ achieves faster and more focused reasoning improvement.

\item \textbf{Data efficiency via verifier-free supervision.} 
SFT relies on ``generate-and-verify'' loops that check against ground-truth answers, posing a major limitation. \model\ avoids this by using the Amateur as a contrastive baseline, turning relative performance gaps into supervision signals. This design removes the dependency on ground-truth verification and extends to domains where definitive solutions are unavailable. By decoupling learning from outcome validation, \model\ focuses on strengthening reasoning processes rather than merely outcomes.

\end{itemize}


\begin{table*}[t]
\centering
\renewcommand{\arraystretch}{1.0}
\setlength{\tabcolsep}{6pt}

\begin{tabular}{lccccccc}
\toprule
\addlinespace[0.1em]
\vcenteredcell{\textbf{Ablation Setting}} &
\vcenteredcell{\textbf{GSM8K}} &
\vcenteredcell{\textbf{MATH}} &
\vcenteredcell{\textbf{SVAMP}} &
\vcenteredcell{\textbf{ASDiv}} &
\vcenteredcell{\textbf{\shortstack{Minerva\\Math}}} &
\vcenteredcell{\textbf{\shortstack{Olympiad\\Bench}}} &
\vcenteredcell{\textbf{AVG.}} \\
\addlinespace[0.1em]
\midrule
\textit{\textit{Qwen2.5-Math-1.5B}}  & 42.5 & 34.2 & 68.8 & 68.1 & 9.9 & 23.7 & 41.2\\
\hdashline
+ Rejection SFT     & 69.2 & 57.1 & 64.1 & 70.2 & 15.1 & 27.6 & 50.6\\
+ GT Supervision      & 43.4 & 34.8 & 70.4 & 69.7 & 10.2 & 19.8 & 41.4\\
\hdashline
\rowcolor{lightpurple} \textbf{+ Full \model} & \textbf{70.6} & \textbf{59.3} & 76.0 & 79.8 & 11.4 & 27.1 & \textbf{54.0}\\
\textcolor{red}{\ding{55}} Select \textcolor{green!60!black}{\ding{51}} Contrast    & 67.6 & 58.8 & 78.7 & 80.5 & 11.0 & 26.4 & 53.8\\
\textcolor{green!60!black}{\ding{51}} Select \textcolor{red}{\ding{55}} Contrast     & 62.0 & 53.1 & 56.6 & 61.0 & 10.7 & 25.5 & 44.8\\
\textcolor{red}{\ding{55}} Select \textcolor{red}{\ding{55}} Contrast      & 55.5 & 50.2 & 50.0 & 65.4 & 10.4 & 24.0 & 42.6\\
\bottomrule
\end{tabular}

\caption{\textbf{Ablation study} on the \model\ framework. We progressively remove key components, \textbf{step selection} and \textbf{contrastive supervision}, to isolate their contributions.}

\label{tab:ablation_results}
\end{table*}

\subsection{Domain Expertise Drives Contrast (RQ3)}
\label{sec:pairing_effects}

Prior contrastive methods rely on rigid parameter-size disparities to create Expert–Amateur contrast. For instance, OPT-13B vs. OPT-125M in Contrastive Decoding (CD)~\citep{li2022contrastive} and LLaMA-65B vs. LLaMA-1.5B in its follow-up study~\citep{o2023contrastive}. Such setups restrict applicability and introducing substantial computational demands. \hyperref[tab:cd_vs_lightreasoner]{Table~\ref*{tab:cd_vs_lightreasoner}} summarizes the key differences between contrastive decoding methods and \model. To overcome this limitation, we hypothesize that domain-specific expertise provides a more applicable axis of contrast than the raw model scale. To validate this hypothesis, we fixed the Expert model and systematically varied the Amateur, progressively narrowing their expertise gap. \hyperref[tab:amateur_variation]{Table~\ref*{tab:amateur_variation}} and \hyperref[fig:gap_vs_perf]{Figure~\ref*{fig:gap_vs_perf}} demonstrate how Expert–Amateur expertise relationships determine \model's effectiveness and offers valuable insights into optimal model contrast pairings.

\begin{itemize}[leftmargin=*]

\item \textbf{Domain expertise as contrast axis.} 
Our results demonstrate that domain-specific knowledge, rather than parameter count, is the primary driver of effective contrastive supervision. The clearest evidence comes from pairing the Qwen2.5-Math-1.5B expert with the generalist Qwen2.5-1.5B amateur, which yields striking performance gains (+12.1\% average gain) despite identical model sizes. This finding frees \model\ from rigid scale requirements and extends it to a broader range of models.

\item \textbf{Effectiveness depends on the expertise gap.} 
As illustrated in \hyperref[fig:gap_vs_perf]{Figure~\ref*{fig:gap_vs_perf}}, performance gains are closely correlated with the size of the expertise gap. 
When Amateur capabilities approach those of the Expert, contrastive signals weaken and benefits diminish. In the extreme case of pairing the Math-1.5B or Math-7B Expert with the stronger Math-1.5B-Instruct model—where the expertise gap is effectively negative—fine-tuning yields negligible gains or even degradation, further proving that expertise differentials are essential for model contrastive learning.

\end{itemize}

\subsection{Ablation Study (RQ4)}
\label{sec:ablation}

To assess the contribution of each core component in the \model\ framework, we conducted a systematic ablation study. By progressively removing individual mechanisms and measuring the impacts, we isolate how each design contributes to overall performance. The results, presented in \hyperref[tab:ablation_results]{Table~\ref*{tab:ablation_results}}, highlight the role of every component and provide key insights into the design of \model. We summarize the findings below:

\begin{figure}[t]
    \centering
    \includegraphics[width=0.7\linewidth]{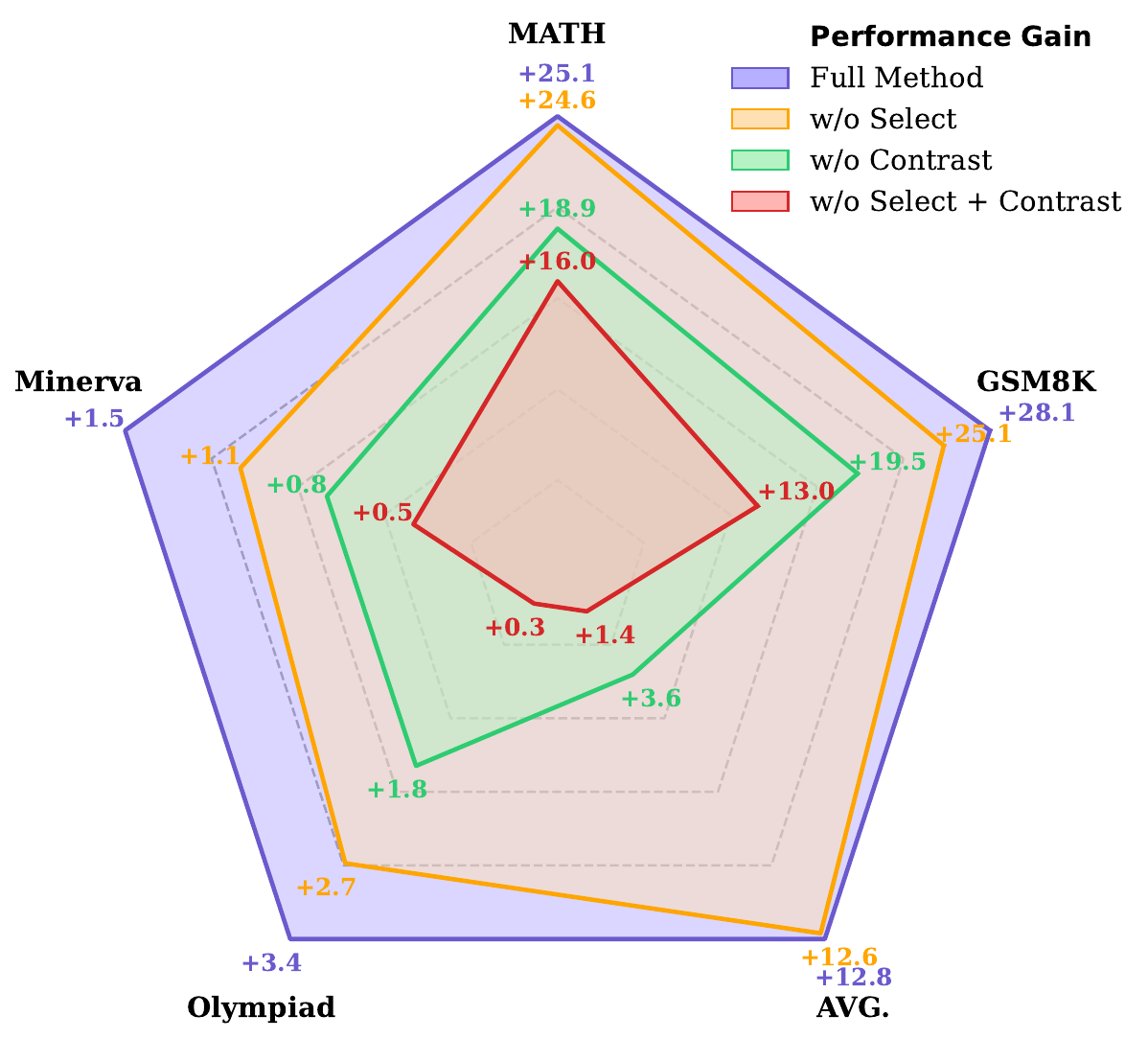}

    \caption{\textbf{Impact of ablation.} Removing key components from \model\ consistently degrades performance, emphasizing their critical roles.}

    \label{fig:ablation_radar}
\end{figure}

\begin{itemize}[leftmargin=*]

\item \textbf{Impact of Informative Step Selection.}  
Removing step selection turns \model\ into a full-sampling pipeline without KL-based pruning. This ablation led to a clear performance decline (e.g., –3.0\% on GSM8K), indicating that many steps contribute noise rather than meaningful learning value. The step-selection filter addresses this by discarding trivial cases, enabling more targeted training.

\item \textbf{Impact of Contrastive Supervision.}  
Removing contrastive supervision reduces \model\ to fine-tuning the Expert on its own paths (filtered by step selection). Without the Amateur’s contrast to capture the Expert’s relative margin, average performance drops by 9.2\%. This demonstrates the central role of contrastive supervision in amplifying the Expert’s strengths while steering it away from amateur-like tendencies.

\item \textbf{Synergy Between Contrast and Selection.}  
When both mechanisms are removed, average performance drops by a staggering 12.4\%—even greater than the sum of the individual ablations (–9.2\% and –0.2\%). This superadditive decline reveals their mutual dependence: without step selection, the contrastive signal is diluted by trivial steps; without contrastive supervision, the high-value steps cannot be transformed into effective learning signals. Together, these mechanisms form a tightly coupled system, explaining the amplified benefits when used jointly.

\item \textbf{Insights from Competitive Approaches.}  
To further validate our method, we compared \model\ against alternative strategies (\hyperref[tab:ablation_results]{Table~\ref*{tab:ablation_results}}). Fine-tuning on human-curated solutions (\textit{GT Supervision}) yielded weak results, while SFT on correct self-generated trajectories provided clear gains but still lagged behind the ablation variant with contrastive supervision alone. These findings highlight a key principle: pretrained models learn most effectively from signals grounded in their own behavior. By selectively amplifying the Expert’s advantage on the most informative reasoning steps, \model\ achieves faster and more robust improvements than either human-curated supervision or rejection-sampled SFT.

\end{itemize}

\section{Conclusion}

In this work, we introduced \textbf{\model}, a novel framework for advancing LLM reasoning by exploiting the behavioral divergence between expert and amateur models. Grounded in the insight that critical reasoning steps carry disproportionate learning value, \model\ integrates targeted step selection with contrastive supervision to amplify the expert’s strengths while minimizing resource demands and training complexity. Experiments across diverse benchmarks demonstrate that \model\ not only outperforms traditional SFT under equal settings, but also delivers order-of-magnitude efficiency and eliminates reliance on labeled ground truth. These results highlight \model\ as a practical efficient solution for building more capable reasoning models.

\section*{Limitations}

While \model\ demonstrates strong performance, several limitations remain:

\begin{itemize}
    \item[(1)] \textbf{Scope of evaluation.} 
    Our experiments focus primarily on mathematical reasoning benchmarks (e.g., GSM8K, MATH, Minerva Math), leaving the generality of the framework to other domains, such as code reasoning, for future exploration.

    \item[(2)] \textbf{Pairing strategy.} 
    The Expert–Amateur contrast relies on model pairs with a balanced capability gap; developing more adaptive or data-driven pairing strategies could yield stronger and more stable supervision signals.

    \item[(3)] \textbf{Hyperparameter sensitivity.} 
    Although the proposed step-selection and contrastive supervision mechanisms effectively reduce resource consumption while maintaining competitive performance, they introduce additional hyperparameters (e.g., $\alpha$-pruning, $\beta$-filtering) that may require careful tuning across different tasks.

    \item[(4)] \textbf{Model scalability.} 
    Our experiments are conducted on small and mid-scale open-weight models; extending the approach to larger proprietary models would further demonstrate its scalability and practical applicability.
    
\end{itemize}

\bibliography{reference}

\appendix
\clearpage
\appendix

\vspace*{2em}
\begin{center}
\LARGE \textbf{Appendix}
\end{center}
\vspace{1.5em}

\section{Related Work}
\label{app:related}

\subsection{Contrastive Decoding}

Contrastive decoding (CD) has emerged as a novel technique for improving inference-time decoding in language models. CD introduces a smaller ``amateur'' model alongside a larger ``expert'' model from the same family, and contrasts their next-token distributions at each decoding step \citep{li2022contrastive}. Instead of directly sampling from the expert’s distribution, CD selects tokens based on the difference between the expert and amateur log-probabilities, effectively reweighting the expert’s predictions. The key intuition is that the amateur model serves as a proxy for undesirable behaviors (e.g., repetition or incoherence). By subtracting these tendencies, CD isolates the ``net expertise'' of the expert model, yielding a more reliable decoding signal. Empirically, this contrastive objective reduces hallucination and improves coherence in open-ended generation.

Subsequent work extended CD beyond open-ended generation to reasoning tasks, where it promotes less noisy token predictions, leading to improved reasoning performance \citep{o2023contrastive}. To reduce the overhead of maintaining two separate models, \citet{phan2024distillation} propose constructing the amateur model by applying distillation techniques (e.g., dropout or quantization) to the expert, enabling a single model to play both roles. More recently, \citet{chang2024explaining} provide a theoretical perspective, interpreting the contrastive scores as approximations to an idealized target distribution.

Despite their promise, CD approaches are subject to several inherent limitations. First, CD incurs substantial computational overhead at inference time, as it requires evaluating both the expert and amateur models at every decoding step. This dual-model dependency increases memory usage and introduces additional latency. Furthermore, CD relies on a sufficiently large capability gap between the expert and amateur models to produce meaningful contrast. In practice, this often necessitates pairing models with a significant size disparity within the same model family, which imposes restrictive design constraints and limits the flexibility of CD.

\model\ builds upon the core intuition of contrastive decoding while introducing a fundamental paradigm shift. Instead of applying contrast at inference time, \model\ leverages expert–amateur divergence during training, distilling it into persistent supervision signals that eliminates concurrent model execution at inference. This removes the memory and latency overhead associated with CD. Moreover, by exploiting domain-specific capability gaps rather than relying solely on model size differences, \model\ relaxes the strict size disparity requirements of traditional CD setups. Altogether, \model\ extends the contrastive paradigm beyond decoding, enabling efficient and robust improvements in reasoning performance.

\subsection{Training on Selective Tokens}

Conventional training paradigms for language models uniformly optimize over all tokens in a trajectory, implicitly treating them as equally informative. Recent work has begun to challenge this convention, recognizing that different tokens carry varying learning value. In this view, uniformly training on all tokens may dilute learning signals by allocating substantial optimization effort to ``low-return'' tokens, while limiting the impact of more informative ``high-return'' tokens \citep{wang2025beyond, sun2026well, meng2026sparse}. Focusing on a subset of informative tokens can improve training efficiency by reducing redundant updates, while potentially enhancing robustness by filtering out noise and trivial patterns.

A central challenge, however, lies in identifying such informative tokens \emph{a priori}. Existing approaches broadly fall into two categories. The first adopts simple, heuristic-based selection strategies that focus on prefix segments of reasoning trajectories, motivated by the observation that early-stage reasoning often shapes downstream outcomes. For example, UPFT \citep{ji2025first} performs SFT exclusively on initial prefixes, while PPPO \citep{sun2026well} restricts policy optimization to early-stage reasoning tokens. The second category employs adaptive, data-driven criteria to identify high-value tokens. RHO-1 \citep{lin2024rho} analyzes token-level training dynamics and trains a reference model to score token importance, selecting only tokens with higher learning utility for large-scale pretraining. Similarly, \citet{wang2025beyond} observe an 80/20 rule, where a small subset of high-entropy tokens drives most of the learning signal; restricting updates to these tokens yields comparable or improved performance over full-trajectory training. 

\model\ introduces a fundamentally different perspective on selective token training. Rather than relying on heuristic truncation or external scoring mechanisms, it leverages the domain-specific expertise gap between an expert and an amateur model within the same family. This capability gap induces systematic differences in their next-token distributions, where the resulting divergence signals distinctive reasoning behavior exhibited by the expert. These high-divergence tokens naturally correspond to steps where the expert demonstrates reasoning patterns absent in the amateur.

\subsection{Self-Distillation}

Knowledge distillation \citep{hinton2015distilling} is a widely used technique for improving language model performance by transferring knowledge from a teacher model to a student model. It offers several practical advantages, including token-level supervision, algorithmic simplicity, and stable optimization. To reduce reliance on external teacher models, self-distillation has been proposed as an extension of this paradigm, where a model improves by learning from targets derived from its own generations. In this setting, the model first produces candidate outputs, which are then filtered or refined to construct supervision signals for further training. Rejection sampling fine-tuning \citep{yuan2023scalingrelationshiplearningmathematical}, widely adopted as post-training SFT, can be viewed as a form of self-distillation, where the model is trained to imitate higher-quality versions of its own generated trajectories.

Recently, self-distillation has gained renewed attention with the emergence of on-policy self-distillation (OPSD)\footnote{OPSD emerged after the initial version of our work; we include it in this revised version for completeness and discuss its relation to our method.}. Building on the framework of \citet{snell2022learning}, several concurrent works explore this paradigm. SDFT \citep{shenfeld2026self} and Self-Distilled Reasoner \citep{zhao2026self} propose using a single model as both teacher and student by conditioning on different contexts, where the teacher has access to privileged information unavailable to the student. SDPO \citep{hubotter2026reinforcement} further extends this framework by incorporating additional feedback signals, such as execution errors or evaluation outcomes, into the privileged context. Despite differences in implementation, these methods share a common principle: leveraging the model's in-context learning ability to construct a stronger target from enriched inputs.

\model\ adopts a self-distillation perspective while taking a different approach from OPSD-style methods. Instead of relying on auxiliary privileged information to induce a teacher--student asymmetry, \model\ leverages the native capability gap between an expert and an amateur model. Unlike OPSD, which follows a ``strong-to-weak'' paradigm, \model\ does not treat the amateur as a teacher. Rather, the amateur serves as a reference signal that highlights the expert's relative strengths. This avoids the need for artificially constructed signals, which can be brittle when the privileged information is noisy \citep{hubotter2026reinforcement}, and also prevents the train--inference mismatch observed in OPSD \citep{yang2026selfdistilledrlvr}. In addition, \model\ introduces two key mechanisms into the self-distillation framework: selective training and a contrastive objective. The former improves efficiency by focusing updates on high-value tokens, while the latter reduces target noise and improves supervision quality. While \model\ introduces these novel components, it remains largely off-policy. Exploring on-policy extensions of \model\ is an important direction for future work.

\section{The Use of Large Language Models}  
We used OpenAI’s ChatGPT solely as a writing assistant to polish grammar, phrasing, and readability of the paper. All research content is entirely the work of the authors.

\begin{algorithm*}[t]
\caption{\model: An efficient self-distillation framework for reasoning enhancement.}
\label{alg:lr}
\begin{algorithmic}[1]
\Require Expert model $\pi_E$, Amateur model $\pi_A$, dataset $\mathcal{Q}$
\Ensure Enhanced Expert model $\pi'_E$

\Phase{Phase 1: Contrastive Sampling}
\ForAll{problem $q \in \mathcal{Q}$}
  \State Generate a CoT trajectory using $\pi_E$: $\{(s_t, \pi_E(\cdot \mid s_t))\}_{t=1}^T$
  \For{$t = 1$ to $T$}
    \State Compute $\pi_A(\cdot \mid s_t)$
    \State Compute $D_{\mathrm{KL}}\!\left(\pi_E(\cdot \mid s_t) ~\|~ \pi_A(\cdot \mid s_t)\right)$
    \If{$D_{\mathrm{KL}}\!\left(\pi_E(\cdot \mid s_t) ~\|~ \pi_A(\cdot \mid s_t)\right) > \beta$}
      \ForAll{$a_i \in \mathcal{A}$}
        \If{$\pi_E(a_i \mid s_t) \ge \alpha \cdot \max\limits_{a} \pi_E(a \mid s_t)$}
          \State Add $a_i$ to $\mathcal{A}_{\text{mask}}$
          \State Compute $v'_C(a_i \mid s_t) \gets \log \left( \frac{\pi_E(a_i \mid s_t)}{\pi_A(a_i \mid s_t)} \right)$
        \EndIf
      \EndFor
      \State Normalize: $\tilde{v}_C(\cdot \mid s_t) \gets \mathrm{softmax} \left( v'_C(\cdot \mid s_t) \right)$ over $\mathcal{A}_{\text{mask}}$
      \ForAll{$a_i \in \mathcal{A}$}
        \If{$a_i \in \mathcal{A}_{\text{mask}}$}
          \State $v_C(a_i \mid s_t) \gets \tilde{v}_C(a_i \mid s_t)$
        \Else
          \State $v_C(a_i \mid s_t) \gets 0$
        \EndIf
      \EndFor
      \State Store $(s_t, v_C(\cdot \mid s_t))$ for training
    \EndIf
  \EndFor
\EndFor

\Phase{Phase 2: Contrastive Fine-tuning}
\ForAll{stored $(s_t, v_C(\cdot \mid s_t))$}
  \State Compute current output $\pi_E(\cdot \mid s_t)$
  \State Compute loss $\mathcal{L}(s_t) \gets D_{\mathrm{KL}}\!\left(v_C(\cdot \mid s_t) \,\big\|\, \pi_E(\cdot \mid s_t)\right)$
  \State Update $\pi_E$ using gradient $\nabla \mathcal{L}(s_t)$ to obtain the fine-tuned model $\pi'_E$
\EndFor
\end{algorithmic}
\end{algorithm*}

\section{From KL Divergence to Contrast Score}
\label{app:kl2contrast}

In \hyperref[sec:tokeninfo]{\S\ref*{sec:tokeninfo}}, we analyzed Expert–Amateur disagreement via KL divergence and observed that reasoning steps where the Expert holds a clear advantage correspond to higher KL divergence values. A natural idea, therefore, is to train the Expert model to maximize this divergence, which is equivalent to minimizing the loss:
\begin{align}
    \mathcal{L}_{\text{KL}}(s_t) = -D_{\mathrm{KL}}\!\left(\pi_E(\cdot\mid s_t)\,\|\,\pi_A(\cdot\mid s_t)\right). \label{eq:maxkld_heuristic}
\end{align}
Treating $\pi_A$ as fixed, the gradient with respect to the Expert parameters $\theta_E$ is:
\begin{align}
&\nabla_{\theta_E}\,\mathcal{L}_{\text{KL}}(s_t)\\ \nonumber
= &\nabla_{\theta_E}\Big[-D_{\mathrm{KL}}\!\big(\pi_E(\cdot\mid s_t)\,\|\,\pi_A(\cdot\mid s_t)\big)\Big] \\ \nonumber
= &-\,\nabla_{\theta_E}\sum_{a\in\mathcal{A}}\pi_E(a\mid s_t)\,\log\frac{\pi_E(a\mid s_t)}{\pi_A(a\mid s_t)} \\ \nonumber
= &-\sum_{a\in\mathcal{A}}\nabla_{\theta_E}\!\Big[\pi_E(a\mid s_t)\,\log\tfrac{\pi_E(a\mid s_t)}{\pi_A(a\mid s_t)}\Big], 
\end{align}
By \emph{product rule}:
\begin{align}
     &\nabla_{\theta_E}\,\mathcal{L}_{\text{KL}}(s_t)\nonumber\\
     = &-\sum_{a\in\mathcal{A}}\Big[\log\tfrac{\pi_E(a\mid s_t)}{\pi_A(a\mid s_t)}\,\nabla_{\theta_E}\pi_E(a\mid s_t)\nonumber\\
     &+ \pi_E(a\mid s_t)\,\nabla_{\theta_E}\log\tfrac{\pi_E(a\mid s_t)}{\pi_A(a\mid s_t)}\Big],
     \label{eq:product_rule}
\end{align}
Note that by \emph{chain rule}:
\begin{align}
     &\nabla_{\theta_E}\log\frac{\pi_E(a\mid s_t)}{\pi_A(a\mid s_t)}\nonumber\\
     = &\nabla_{\theta_E}\Big[\log\pi_E(a\mid s_t)-\log\pi_A(a\mid s_t)\Big]\nonumber\\
     = &\nabla_{\theta_E}\log\pi_E(a\mid s_t) \nonumber\\
     = &\frac{\nabla_{\theta_E}\pi_E(a\mid s_t)}{\pi_E(a\mid s_t)},
     \label{eq:chain_rule}
\end{align}
Substituting \eqref{eq:chain_rule} into \eqref{eq:product_rule} yields:
\begin{align}
    &\nabla_{\theta_E}\,\mathcal{L}_{\text{KL}}(s_t)\\ \nonumber
    = &-\sum_{a\in\mathcal{A}}\Big(\log\tfrac{\pi_E(a\mid s_t)}{\pi_A(a\mid s_t)}+1\Big)\,\nabla_{\theta_E}\pi_E(a\mid s_t).
\end{align}
At this point, the log-ratio naturally emerges:
\begin{align}
v'_C(a \mid s_t)
&\;=\; \log \frac{\pi_E(a\mid s_t)}{\pi_A(a\mid s_t)} \\ \nonumber
&\;=\; \log \pi_E(a \mid s_t) - \log \pi_A(a \mid s_t), \label{eq:weight_score}
\end{align}
which coincides exactly with the \emph{contrast score} proposed in \citet{li2022contrastive}. This observation motivates our choice: rather than directly maximizing KL divergence, we adopt the contrast score as the central mechanism for constructing the supervision signal to guide the Expert model.

\section{Connection between Selection, Contrast, and Training}
\label{app:connect3}

As shown in \hyperref[app:kl2contrast]{\S\ref*{app:kl2contrast}}, a central quantity in our framework is the \emph{contrast score} \citep{li2022contrastive}:
\begin{align}
    v'_C(a \mid s_t) \;=\; &\log \pi_E(a \mid s_t) \\ \nonumber
    - &\log \pi_A(a \mid s_t),
\end{align}
This log-ratio underlies step selection, contrastive supervision, and the training objective, unifying the three components into a mutually reinforcing and coherent framework.

\paragraph{Step Selection.}
Expert–Amateur disagreement is quantified by the KL divergence:
\begin{align}
&D_{\mathrm{KL}}\!\left(\pi_E(\cdot \mid s_t)\,\|\,\pi_A(\cdot \mid s_t)\right) \\ \nonumber
= &\sum_{a \in \mathcal{A}} \pi_E(a \mid s_t)\, v'_C(a \mid s_t). \\ \nonumber
= & \mathbb{E}_{a\sim \pi_E(\cdot \mid s_t)}[v'_C(a \mid s_t)]
\end{align}

\paragraph{Contrastive Distributional Supervision.}
Restricting to a masked vocabulary $\mathcal{A}_{\text{mask}}$, we normalize $v'_C(a \mid s_t), a \in \mathcal{A}_{\text{mask}}$ via:
\begin{align}
\tilde{v}_C(a \mid s_t) 
= \frac{\exp(v'_C(a \mid s_t))}{\sum_{b \in \mathcal{A}_{\text{mask}}}\exp(v'_C(b \mid s_t))},
\end{align}
and extend to the full vocabulary $a \in \mathcal{A}$ by:
\begin{align}
    v_C(a \mid s_t) 
    &= \tilde{v}_C(a \mid s_t) \cdot \mathbbm{1}\!\left[ a \in \mathcal{A}_{\text{mask}} \right], 
\end{align}

\paragraph{Training Objective.}
The Expert is distilled toward $v_C(a \mid s_t) $ by minimizing forward KL:
\begin{align}
\mathcal{L}(s_t) 
= &D_{\mathrm{KL}}\!\left(v_C(\cdot \mid s_t)\,\|\,\pi_E(\cdot \mid s_t)\right) \\ \nonumber
= &\sum_{a \in \mathcal{A}} v_C(a \mid s_t)\,\log \frac{v_C(a \mid s_t)}{\pi_E(a \mid s_t)}.
\end{align}

\paragraph{Unification.}
All three components are governed by the same log-ratio $v'_C(a \mid s_t)$: 
\begin{enumerate}[leftmargin=*]
    \item \textbf{Step selection} quantifies divergence through expected value of $v'_C(a \mid s_t)$ and retains informative reasoning steps.
    \item \textbf{Contrastive supervision} transforms $v'_C(a \mid s_t)$ into a learning signal, capturing the Expert model’s difference margin from the Amateur.
    \item \textbf{Training} propagates gradients implicitly weighted by $v'_C(a \mid s_t)$, guiding toward high-contrast tokens.  
\end{enumerate}

\paragraph{Intuitive explanation.}

The expert and amateur models are drawn from the same model family (e.g., the expert is Qwen2.5-Math while the amateur is Qwen2.5), sharing identical architectures and pretraining. The expert can be viewed as the amateur model after additional domain-specific post-training. As a result, the two models exhibit similar syntactic and low-level statistical structure, and their divergence primarily reflects differences in domain expertise. In particular, this divergence tends to concentrate on directions where the expert demonstrates stronger reasoning ability shaped by its additional training.

From this perspective, the difference between the two policies can be interpreted as a noisy estimate of a performance improvement direction. It highlights regions where the expert assigns higher probability than the amateur, which often correspond to more accurate reasoning steps. The noise arises because not all differences reflect genuine improvements, but it is largely unstructured and tends to average out.

The KL divergence $D_{\mathrm{KL}}\!\left(\pi_E(\cdot\mid s)\,\|\,\pi_A(\cdot\mid s)\right)$ provides a natural mechanism to aggregate this signal. Its gradient emphasizes directions where the expert deviates from the amateur, and serves as a good estimation of the performance improvement direction. Consequently, maximizing this divergence encourages updates that further reinforce regions where $\pi_E$ outperforms $\pi_A$, while placing less emphasis on regions where the two models agree or where the signal is weak.

Although individual differences may be noisy, repeated sampling combined with filtering acts as a form of variance reduction. In expectation, this process nudges updates toward consistent advantageous directions. Intuitively, the method reinforces regions where the expert is reliably stronger, rather than indiscriminately shifting the entire distribution, thereby helping to avoid pathological behavior collapse.

\section{Relation to Reinforcement Learning}
\label{app:relate2rl}

\paragraph{Policy gradient methods.}
In reinforcement learning (RL), the objective of policy gradient methods is to maximize the expected return
\begin{align}
    J(\theta) \;=\; \mathbb{E}_{\pi_\theta}\!\left[ \sum_{t=0}^{T-1} \gamma^{t}\, r_t \right],
\end{align}
where $r_t$ is the reward at step $t$ and $\gamma \in [0,1]$ is a discount factor. Under the actor–critic framework, the policy gradient theorem states
\begin{align}
&\nabla_\theta J(\theta) \nonumber \\
&= \mathbb{E}_{\pi_\theta}\!\left[
    \sum_{t=0}^{T-1} 
    \nabla_\theta \log \pi_\theta(a_t \mid s_t)\, 
    A^\theta(a_t, s_t)\right]
\label{eq:policy_grad}
\end{align}
where $A^\theta(a_t, s_t)$ denotes the advantage of action $a_t$ at state $s_t$.

\paragraph{Parallel with our framework.}
In our setting, the training objective is the KL divergence:
\begin{align}
    \mathcal{L}(s_t) = & D_{\mathrm{KL}}\!\big(v_C(\cdot \mid s_t)\,\|\,\pi_E(\cdot \mid s_t)\big) \nonumber \\
    = & \sum_{a \in \mathcal{A}} v_C(a \mid s_t)\, \log \frac{v_C(a \mid s_t)}{\pi_E(a \mid s_t)} \nonumber \\
    = & \sum_{a \in \mathcal{A}} v_C(a \mid s_t)\, \log v_C(a \mid s_t) \nonumber \\
    & - \sum_{a \in \mathcal{A}} v_C(a \mid s_t)\,\log \pi_E(a \mid s_t). \label{eq:kl-appendix}
\end{align}
The first term in~\eqref{eq:kl-appendix} is constant with respect to $\pi_E$. Therefore, the gradient of the loss with respect to $\pi_E$ reduces to:
\begin{align}
    &\nabla_{\theta_E}\mathcal{L}(s_t) \nonumber \\
    &= - \sum_{a \in \mathcal{A}} v_C(a \mid s_t) \,\nabla_{\theta_E}\log \pi_E(a \mid s_t). \label{eq:our_update}
\end{align}
This update has the same structure as a policy gradient step. In particular, it can be written as:
\begin{align}
    &\nabla_{\theta_E}\mathcal{L}(s_t) \\ \nonumber
    &= - \sum_{a \in \mathcal{A}} A^{\text{LightReasoner}}(a, s_t)\,\nabla_{\theta_E}\log \pi_E(a \mid s_t),
\end{align}
where $A^{\text{LightReasoner}}(a, s_t) = v_C(a \mid s_t)$.

\paragraph{Inner connection.}
In standard RL, this weight corresponds to the advantage $A^{\pi_\theta}(s_t, a_t)$. In our framework, the contrastive target $v_C(a \mid s_t)$ plays an analogous role as an advantage signal. The key distinction lies in the optimization direction: policy gradient methods perform gradient ascent to maximize $J(\theta)$, whereas we perform gradient descent to minimize $\mathcal{L}(s_t)$.

Moreover, the contrastive score $v_C(a \mid s_t)$ is a masked and renormalized version of the log-ratio $v'_C(a \mid s_t)$ defined in~\eqref{eq:weight_score}. As a result, the training objective is closely aligned with the heuristic of maximizing the KL divergence between the expert and amateur models in~\eqref{eq:maxkld_heuristic}.

This perspective suggests that our framework can be viewed as a variant of policy gradient training, where the advantage signal arises from Expert--Amateur divergence rather than from environment rewards or human feedback. 

Recent work on on-policy self-distillation (OPSD) \citep{shenfeld2026self,
hubotter2026reinforcement} similarly explores deriving advantage-like signals from the model itself, for instance through log-ratio-based comparisons under different conditioning contexts. Exploring how \model\ can be extended into a fully on-policy framework with explicit policy updates and reward signals remains an interesting direction for future work.

\section{Entropy Dynamics}
\label{app:entropy}

Policy updates instantiate an exploration--exploitation trade-off. Recent work shows that this trade-off is closely reflected in \emph{policy entropy} \citep{cui2025entropy}: without explicit control, entropy often collapses early, leading to overconfident policies and premature performance saturation. Empirically, pretrained models with higher initial entropy tend to achieve better downstream performance after RL, while the RL process itself largely converts entropy into reward \citep{yue2025rl}. This makes entropy preservation an important consideration in LLM post-training.

\paragraph{Policy entropy.}
Formally, the entropy of a policy $\pi$ relative to a dataset $\mathcal{D}$ is
\begin{align}
    & H(\pi,\mathcal{D}) \\ 
    & =
    -\,\mathbb{E}_{\mathbf{y}_{<t} \sim \mathcal{D}}
    \Bigg[
    \sum_{y_t} \pi(y_t \mid \mathbf{y}_{<t}) \log \pi(y_t \mid \mathbf{y}_{<t})
    \Bigg]. \nonumber
\end{align}

\paragraph{Entropy change under policy gradient.}
For a single state $s$, the intrinsic policy entropy is
\begin{align}
    H(\pi \mid s) := - \sum_{a} \pi(a\mid s)\,\log \pi(a\mid s).
\end{align}
For a tabular softmax policy updated by one step of vanilla policy gradient with step size $\eta$, the stepwise entropy change satisfies the first-order approximation
\begin{align}
    & H(\pi^{k+1}\!\mid s)-H(\pi^{k}\!\mid s)
    \approx -\eta\;\\
    & \operatorname{Cov}_{a\sim \pi^{k}(\cdot\mid s)}
    \Bigl(
        \log \pi^{k}(a\mid s),\,
        \pi^{k}(a\mid s)\,A^{k}(s,a)
    \Bigr), \nonumber
\end{align}
where $A^{k}(s,a)$ is the advantage, satisfying $\mathbb{E}_{a\sim\pi^{k}}[A^{k}(s,a)] = 0$. Thus, when high-probability actions tend to carry positive advantage, the covariance is positive and entropy decreases; when advantage is concentrated on low-probability actions, the covariance can become negative and entropy can increase \citep{cui2025entropy}.

\paragraph{Entropy change with contrast score.}
As established in \hyperref[app:relate2rl]{\S\ref*{app:relate2rl}}, our framework has the same update structure as policy gradient when the contrast score is interpreted as the advantage signal. Consequently, the first-order change in policy entropy can be written as
\begin{align}
  &H(\pi_E^{k+1}\!\mid s)-H(\pi_E^{k}\!\mid s)
  \approx -\,\eta\, \label{eq:entropy-contrast-start} \\
  &\operatorname{Cov}_{a\sim \pi_E^{k}(\cdot\mid s)}
  \Bigl(
      \log \pi_E^{k}(a\mid s),\,
      \pi_E^{k}(a\mid s)\,v_C^{k}(s,a)
  \Bigr). \nonumber
\end{align}
Let $\mathcal{A}_{\mathrm{mask}}(s)\subseteq\mathcal{A}$ denote the masked action set, and let $\mathbbm{1}[a \in \mathcal{A}_{\mathrm{mask}}(s)]$ be its indicator function. Define the masked, temperature-scaled contrast as
\begin{equation}
\begin{aligned}
  v_C^{k}(s,a)
  &:=
  \frac{1}{\tau}\,\mathbbm{1}[a \in \mathcal{A}_{\mathrm{mask}}(s)]\,v_C^{\prime\,k}(s,a), \\
  v_C^{\prime\,k}(s,a)
  &:=
  \log\frac{\pi_E^{k}(a\mid s)}{\pi_A^{k}(a\mid s)},
\end{aligned}
\label{eq:entropy-contrast-middle}
\end{equation}
where $\tau > 0$ is a temperature scaling parameter. Substituting \eqref{eq:entropy-contrast-middle} into \eqref{eq:entropy-contrast-start} yields
\begin{align}
  & H(\pi_E^{k+1}\!\mid s)-H(\pi_E^{k}\!\mid s)
  \approx \label{eq:entropy-contrast-final} \\
  & -\,\frac{\eta}{\tau}\,
  \operatorname{Cov}_{a\sim \pi_E^{k}(\cdot\mid s)}
  \Bigl(
      \log \pi_E^{k}(a\mid s),\, \nonumber \\
  & \pi_E^{k}(a\mid s)\, 
      \mathbbm{1}[a \in \mathcal{A}_{\mathrm{mask}}(s)]\,
      \log\frac{\pi_E^{k}(a\mid s)}{\pi_A^{k}(a\mid s)}
  \Bigr). \nonumber
\end{align}
Intuitively, \eqref{eq:entropy-contrast-final} shows that the entropy change is driven by terms of the form
\begin{align}
    \pi_E^{k}(a\mid s)\,
    \mathbbm{1}[a \in \mathcal{A}_{\mathrm{mask}}(s)]\,
    \log\frac{\pi_E^{k}(a\mid s)}{\pi_A^{k}(a\mid s)}.
\end{align}
The $\alpha$-mask restricts updates to actions with non-negligible probability under the Expert. If the Expert and Amateur assign similar probabilities to an action, then the log-ratio is close to zero, and the corresponding contribution to entropy change is negligible even when $\pi_E$ is large. In contrast, when the Expert places substantial probability mass on actions for which it strongly outperforms the Amateur, the log-ratio becomes large and positive, making the covariance term positive and causing entropy to decrease more substantially. Therefore, the contrastive signal spends entropy selectively on high-value, high-contrast actions rather than on regions where the two models already agree. This makes entropy reduction more targeted, helping avoid premature entropy collapse while preserving the Expert's exploration capacity in low-signal regions.

\begin{table*}[t]
\centering
\small
\setlength{\tabcolsep}{6pt}
\begin{tabular}{lcccccc}
\toprule
\textbf{Sampling Length} & \textbf{GSM8K} & \textbf{MATH} & \textbf{SVAMP} & \textbf{ASDiv} & \textbf{Minerva} & \textbf{Olympiad} \\
\midrule
64  & \textbf{74.3} & 61.5 & \textbf{81.5} & \textbf{81.4} & \textbf{18.8} & 24.6 \\
128 & 70.6 & 59.3 & 76.0 & 79.8 & 11.4 & \textbf{27.1} \\
256 & 72.3 & \textbf{62.2} & 78.5 & 79.4 & 18.4 & 21.3 \\
\bottomrule
\end{tabular}

\caption{Performance under different sampling rollout lengths. Results do not exhibit a monotonic trend, suggesting that \model\ is largely insensitive to this parameter.}

\label{tab:truncation}
\end{table*}

\begin{table*}[t]
\centering
\small
\renewcommand{\arraystretch}{1.1}
\setlength{\tabcolsep}{6pt}
\begin{tabular}{lccc}
\toprule
\textbf{Model} & \textbf{CommonsenseQA} & \textbf{HellaSwag} & \textbf{ARC-Challenge} \\
\midrule
Baseline & 62.6 & 46.2 & 42.8 \\
+ \model & \textbf{64.2} & \textbf{47.4} & \textbf{45.6} \\
\bottomrule
\end{tabular}

\caption{Generalization performance on non-math reasoning benchmarks. Despite training solely on GSM8K, \model\ demonstrates effective transfer beyond the training domain.}

\label{tab:generalization}
\end{table*}

\section{Additional Experiments}
\label{app:additionalexp}

\subsection{Effect of Truncation Length}
\label{app:truncation}

In the \model\ framework, supervision examples are constructed from short sampling rollouts (e.g., the first 128 tokens of the model's reasoning trajectory). Importantly, generation is paused once the maximum token limit is reached, rather than forcing the model to compress full solutions into shorter forms. Each supervision example is constructed at the next-token level, which avoids introducing length-dependent biases or fragmented reasoning patterns.

To verify that the model is not biased toward shallow reasoning, we include qualitative case studies in \hyperref[app:casestudy]{Appendix~\ref*{app:casestudy}}, showing that \model-trained models produce full, coherent CoT solutions.

We further conduct a controlled experiment by varying the sampling rollout length (i.e., the number of generated tokens before pausing) during supervision construction, as shown in \hyperref[tab:truncation]{Table~\ref*{tab:truncation}}. Across benchmarks, performance does not follow a monotonic trend with respect to the rollout length, with different tasks favoring different prefix lengths. This suggests that \model\ is not sensitive to the exact sampling rollout length, and that performance gains primarily stem from the contrastive signal derived from informative steps rather than from length-dependent factors.

\subsection{Generalization Beyond Math}
\label{app:generalization}

To evaluate whether our method transfers beyond the math domain, we evaluate the base model Qwen2.5-Math-1.5B and its \model-trained counterpart on three standard commonsense reasoning benchmarks. The results are shown in \hyperref[tab:generalization]{Table~\ref*{tab:generalization}}. 

Although training is conducted solely on supervision samples derived from GSM8K, a math dataset, \model\ consistently improves performance across all three commonsense benchmarks. This suggests that our method captures and enhances fundamental reasoning capabilities that generalize beyond the training domain.

\section{Supplementary Details}
\label{app:supplementary}

We provide additional details to complement the descriptions in 
\hyperref[sec:model_data]{\S\ref*{sec:model_data}} and 
\hyperref[sec:training_config]{\S\ref*{sec:training_config}}, covering both our proposed method, \model, and the competitive supervised fine-tuning (SFT) baseline.

\subsection{Datasets}
\label{app:datasets}

We exclusively use the \textbf{GSM8K} \citep{cobbe2021training} training set, a collection of grade-school math problems emphasizing step-by-step reasoning, to generate contrastive samples. To evaluate the transferability of the learned skills, we assess our models on a diverse suite of benchmarks: \textbf{MATH} \citep{hendrycks2021measuring}, a collection of high school competition problems; \textbf{SVAMP} \citep{patel2021nlp} and \textbf{ASDiv} \citep{miao2021diverse}, testing numerical reasoning through linguistically varied arithmetic problems; \textbf{Minerva Math} \citep{lewkowycz2022solving}, quantitative problems from advanced STEM courses; \textbf{OlympiadBench} \citep{he2024olympiadbench}, challenging problems from international math olympiads; and \textbf{MMLU-STEM} \citep{hendrycks2020measuring}, which evaluates broad knowledge across science and math. This range spans from foundational arithmetic to expert-level reasoning, enabling a thorough assessment of both generalization and specialization.

\subsection{Baseline Models}
\label{app:models}

Our method leverages pairing between \textit{Expert} and \textit{Amateur} models to generate training signals:
\begin{itemize}[leftmargin=*]
    \item \emph{\textbf{Expert Models}} span varying capabilities and sizes to ensure robust evaluation:
    \textbf{(1) Qwen2.5-Math-1.5B} and \textbf{(2) Qwen2.5-Math-7B} are derived from Qwen2.5 base models via pretraining on a 1T-token math corpus \citep{yang2024qwen2}.  
    \textbf{(3) Qwen2.5-Math-1.5B-Instruct} and \textbf{(4) Qwen2.5-Math-7B-Instruct} receive additional multi-stage post-training, including supervised fine-tuning and GRPO-based reinforcement learning \citep{yang2024qwen2}.   
    \textbf{(5) DeepSeek-R1-Distill-Qwen-1.5B} is fine-tuned based on the corresponding Qwen2.5-Math model, using teacher-curated examples generated by DeepSeek-R1 \citep{guo2025deepseek}.

    \item \emph{\textbf{Amateur Model}} is fixed as \textbf{Qwen2.5-0.5B}, a base model without specialized mathematical pretraining but with general linguistic reasoning ability \citep{yang2024qwen2}. Being in the same model family as the Experts ensures that performance differences reflect mathematical expertise rather than architectural discrepancies.
\end{itemize}

\subsection{\model}
\label{app:lr_extra}

For \model, hyperparameters were selected to ensure training efficiency and final performance. Below we report further details of the experimental setup.

\subsubsection{$\alpha$-masking}
Following \citet{li2022contrastive}, we apply $\alpha$-masking for $a \in \mathcal{A}$ to truncate the Expert distribution:
\begin{align}
    \mathcal{A}_{\text{mask}} 
    = \left\{\pi_E(a \mid s_t) \ge \alpha \cdot \max_{b \in \mathcal{A}} \pi_E(b \mid s_t) \right\}.
\end{align}
This operation trades off quality and diversity: larger $\alpha$ keeps only top-probability tokens, yielding higher quality but reduced coverage, while smaller $\alpha$ admits more diverse candidates at the cost of reliability. In our framework, the Expert and Amateur share the same vocabulary and tokenizer, and $\alpha \in [0,1]$ controls how aggressively the Expert’s next-token distribution is truncated. Tokens below the threshold are excluded, and contrastive supervision is applied only over the surviving set.  

We set $\alpha=0.2$ throughout, deviating from the $\alpha=0.1$ commonly used in \citet{li2022contrastive} and \citet{o2023contrastive}. This choice was motivated by manual inspection across five different models: we compared $\alpha \in \{0.1,0.2,0.4\}$ on a variety of examples and found that $\alpha=0.2$ strikes the best balance. It preserves high-quality candidates while avoiding distributional collapse into a near one-hot target, which can occur under more aggressive $\alpha$-truncation.

\subsubsection{$\beta$-filtering}
As discussed in \hyperref[sec:tokeninfo]{\S\ref*{sec:tokeninfo}} and \hyperref[sec:step_selection]{\S\ref*{sec:step_selection}}, not all tokens in a reasoning trajectory are equally informative. We proposed $\beta$-filtering to address this, retaining only those steps where the Expert–Amateur divergence exceeds a threshold:
\begin{align}
    D_{\mathrm{KL}}\!\big(\pi_E(\cdot \mid s_t)\,\|\,\pi_A(\cdot \mid s_t)\big) > \beta.
\end{align}

\noindent
Here, we provide additional insight into the mechanism of $\beta$-filtering. Empirically, we observe two types of contrastive supervision signals:
\begin{itemize}[leftmargin=*]
    \item \textbf{Single-token labels:} When the Expert’s distribution is sharply peaked, $\alpha$-masking retains only the top-1 token, yielding a degenerate label of $[1.0]$.
    \item \textbf{Multi-token labels:} When the Expert’s distribution is more spread out, multiple tokens survive $\alpha$-masking, producing a distributed label.
\end{itemize}
In practice, $\beta$ effectively regulates the balance between these two cases. Single-token labels almost always arise from \emph{low-KL} steps: a highly peaked Expert distribution indicates strong confidence, which typically occurs on easier steps where Expert and Amateur agree, resulting in low divergence. Such cases contribute little useful contrast. 

Our ablation confirms this effect: removing $\beta$ increases the proportion of single-token labels by $+35.6\%$ for Qwen2.5-Math-1.5B, $+33.5\%$ for Qwen2.5-Math-7B, and $+27.4\%$ for Qwen2.5-Math-1.5B-Instruct. $\beta$-filtering governs the proportion of contrast-rich samples and ensures that fine-tuning focuses on genuine Expert–Amateur disagreements. We fixed $\beta=0.4$ throughout, chosen via manual inspection of the collected examples, and leave the exploration of potentially more optimal choices to future work.

\subsubsection{Prompting}
\label{app:lr_prompting}

Recall the two central stages of \model:
\begin{itemize}[leftmargin=*]
    \item \textbf{Sampling.} For a given problem $q$, we prompt the Expert $\pi_E$ to generate a reasoning trajectory and record the sequence of prefix–distribution pairs:
    \begin{align}
        \mathcal{T}_E \;=\; \big\{ \big(s_t,\; \pi_E(\cdot \mid s_t)\big) \big\}_{t=1}^T. \label{eq:pair_sequence}
    \end{align}
    For each prefix $s_t$ in \eqref{eq:pair_sequence}, we query the Amateur $\pi_A$ with the \emph{same} input $(q, s_t)$ to obtain $\pi_A(\cdot \mid s_t)$ and construct the contrastive target $v_C(\cdot \mid s_t)$ from the pair $\big(\pi_E(\cdot \mid s_t),\,\pi_A(\cdot \mid s_t)\big)$.

    \item \textbf{Fine-tuning.} For each supervision example $(s_t,\, v_C(\cdot \mid s_t))$, we re-query the current Expert to obtain $\pi_E(\cdot \mid s_t)$ and minimize the forward KL divergence:
    \begin{align}
        \mathcal{L}(s_t) = D_{\mathrm{KL}}\!\left(v_C(\cdot \mid s_t)\,\|\,\pi_E(\cdot \mid s_t)\right) 
    \end{align}

\end{itemize}

\noindent
Throughout both stages we attach a fixed Chain-of-Thought (CoT) cue,
\begin{align}
&\text{\texttt{Please reason step by step, and put}} \nonumber\\
&\text{\texttt{your final answer within \textbackslash boxed\{\}.}}
\label{eq:qwen25cot}
\end{align}
to every query. Using a single, shared prompt template for \emph{all} calls (Expert/Amateur during sampling and Expert during training) is critical: it ensures that (i) Expert–Amateur differences arise from the models rather than prompt mismatch, and (ii) the context used to form $v_C(\cdot \mid s_t)$ is exactly the context against which the training KL is evaluated. This prompt consistency eliminates confounding from input formatting and yields stable, comparable distributions $\pi_E(\cdot \mid s_t)$ and $\pi_A(\cdot \mid s_t)$ across the pipeline.

\subsubsection{LoRA Configuration}
\label{app:lr_lora}

\begin{figure}[h]
\centering
\begin{minipage}{0.48\textwidth}
    \centering
    \renewcommand{\arraystretch}{1.2}
    \setlength{\tabcolsep}{8pt}
    \begin{tabular}{ll}
    \toprule
    \textbf{LoRA Parameter} & \textbf{Value} \\
    \midrule
    Rank ($r$)          & 8 \\
    Scaling factor ($\alpha$) & 16 \\
    Target modules      & \texttt{q\_proj}, \texttt{v\_proj} \\
    Dropout             & 0.05 \\
    Bias                & None \\
    Task type           & Causal LM \\
    \bottomrule
    \end{tabular}
    \captionof{table}{\textbf{LoRA configuration} used in \model, shared across all models.}
    \label{tab:lr_lora_config}
\end{minipage}
\hfill
\begin{minipage}{0.48\textwidth}
    \centering
    \includegraphics[width=\linewidth]{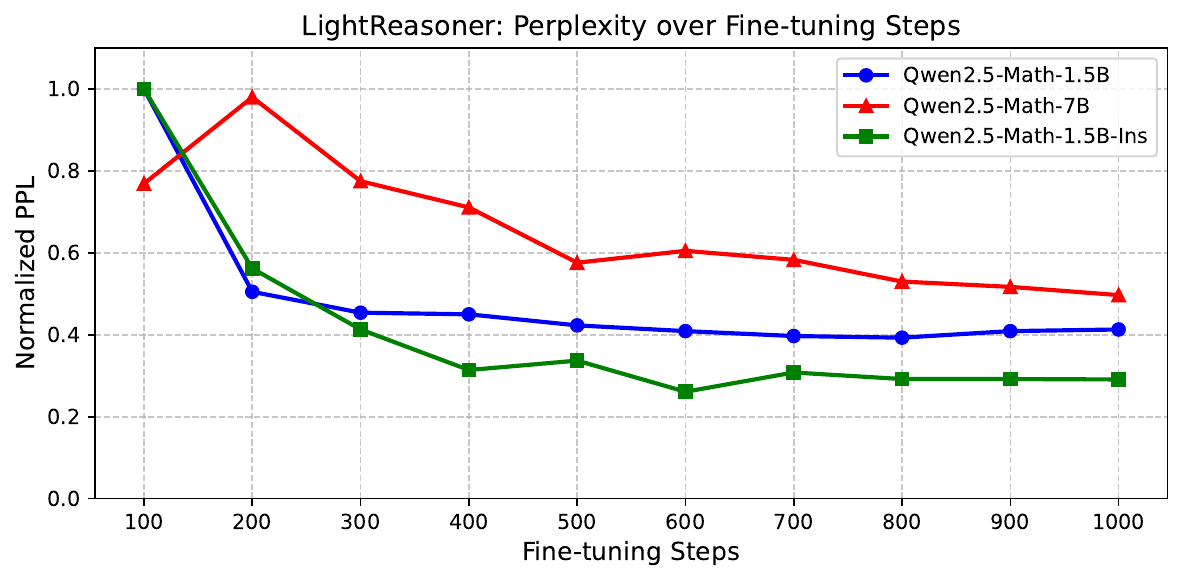}
    \caption{\textbf{Perplexity convergence.} PPL curves show training stabilizes around 1000 steps, supporting our choice of tuning horizon.}
    \label{fig:lr_ppl}
\end{minipage}
\end{figure}

For \model\ fine-tuning, we adopt a LoRA setup \citep{hu2022lora} for efficiency while maintaining stability. The configuration is summarized in \hyperref[tab:lr_lora_config]{Table~\ref*{tab:lr_lora_config}}.

\noindent
Training was performed in \texttt{bfloat16} precision on a single NVIDIA H200 GPU, with the following runtime hyperparameters: batch size of 8 with gradient accumulation of 2 (effective batch size 16), learning rate $5\times 10^{-5}$, and 1000 total update steps. The same configuration was applied across all five backbone models studied in this paper to ensure comparability, while avoiding model-specific hyperparameter tuning. The choice of 1000 steps is further justified by the perplexity curves in \hyperref[fig:lr_ppl]{Figure~\ref*{fig:lr_ppl}}, which show convergence within this horizon for multiple representative models.

\subsection{Supervised Fine-tuning (SFT)}
\label{app:sft_extra}

We provide additional details on the SFT configuration, which serves as the competitive baseline against our method \model.

\subsubsection{Rejection Sampling}
Recent works \citep{yang2024qwen2, guo2025deepseek} commonly employ rejection sampling \citep{yuan2023scalingrelationshiplearningmathematical} in SFT, where models are aligned with demonstrations of correct problem-solving trajectories. This involves generating multiple reasoning traces and retaining only those whose final answers match the ground truth.

In our setting, we adopt a simplified form. For each GSM8K training problem, the base model produces a single reasoning trajectory under the CoT prompt shown in \eqref{eq:qwen25cot}, ensuring fairness and comparability with the \model\ pipeline. We then filter by checking whether the final answer matches the ground truth, retaining only correct trajectories. Depending on the capability of the base model, this yields between 4000 and 7000 problems (see \hyperref[tab:efficiency_comparison]{Table~\ref*{tab:efficiency_comparison}} in \hyperref[sec:efficiency]{\S\ref*{sec:efficiency}}). The resulting set of verified trajectories is used to fine-tune each model.

\subsubsection{Model Fine-tuning}
\begin{figure}[h]
\centering
\begin{minipage}{0.48\textwidth}
    \centering
    \renewcommand{\arraystretch}{1.2}
    \setlength{\tabcolsep}{8pt}
    \begin{tabular}{ll}
    \toprule
    \textbf{LoRA Parameter} & \textbf{Value} \\
    \midrule
    Rank ($r$)          & 8 \\
    Scaling factor ($\alpha$) & 16 \\
    Target modules      & \texttt{q\_proj}, \texttt{v\_proj} \\
    Dropout             & 0.05 \\
    Bias                & None \\
    Task type           & Causal LM \\
    \bottomrule
    \end{tabular}
    \captionof{table}{\textbf{LoRA adapter configuration} used for SFT, applied across all models.}
    \label{tab:sft_lora_config}
\end{minipage}
\hfill
\begin{minipage}{0.48\textwidth}
    \centering
    \includegraphics[width=\linewidth]{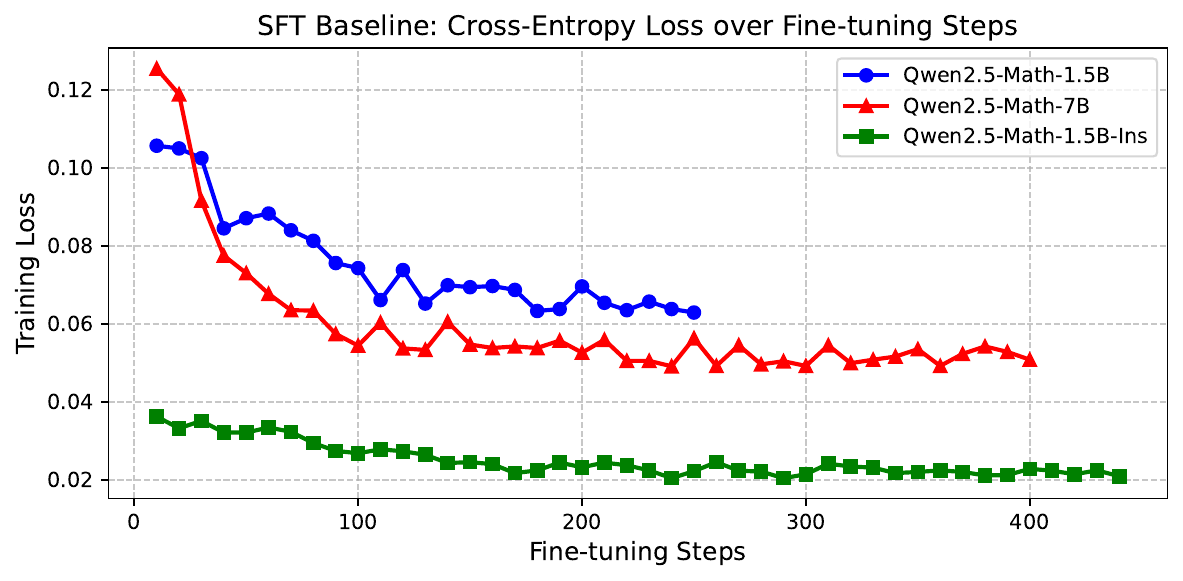}
    \caption{\textbf{SFT training loss.} Curve lengths vary with the number of correct demonstrations, but all runs reach convergence.}
    \label{fig:sft_curve}
\end{minipage}
\end{figure}

In canonical SFT, each training instance consists of a complete reasoning trajectory generated by the base model that concludes with the correct final answer. For each training instance, the model is re-prompted with the fixed CoT prompt shown in \eqref{eq:qwen25cot} along with the problem statement, and is trained under teacher forcing to predict the next token along the gold trajectory until completion. The training objective is the standard cross-entropy loss, computed over the gold trajectory tokens against the model’s predicted distributions. For comparability with \model, we conducted SFT fine-tuning using the same LoRA configuration, as detailed in \hyperref[tab:sft_lora_config]{Table~\ref*{tab:sft_lora_config}}.

\noindent
Training was performed in \texttt{bfloat16} precision on a single NVIDIA H200 GPU, with the following runtime hyperparameters: batch size of 4 with gradient accumulation of 4 (effective batch size 16), learning rate $5\times 10^{-5}$, and a total number of update steps set by the dataset size (e.g., 4K samples correspond to 250 steps). The same configuration was applied to all five backbone models to ensure fair comparison and avoid model-specific tuning. As shown in \hyperref[fig:sft_curve]{Figure~\ref*{fig:sft_curve}}, the training curves vary in length because each model yields a different number of correct demonstrations, but all runs are trained to convergence.

\section{Case Study}
\label{app:casestudy}

Here, we present case studies illustrating how \model\ enhances the base model’s reasoning. Specifically, we show GSM8K test problems where Qwen2.5-Math-1.5B fails but its \model-enhanced counterpart succeeds. Both are evaluated under the zero-shot pass@1 setting:

\begin{itemize}[leftmargin=*]
    \item \textbf{Base model.} We adopt the prompt in \eqref{eq:cot}:
    \begin{align}
        \texttt{"Question: \{input\} Answer: ", "\{output\}"} \label{eq:cot}
    \end{align}
    This choice is motivated by two factors. First, it yields the best base-model accuracy (42.5\% on GSM8K with \eqref{eq:cot} vs. 38.4\% with \eqref{eq:qwen25cot}). Second, under \eqref{eq:qwen25cot}, the base model often degenerates into repeated fragments of the question and partial solutions until the output limit, producing corrupted traces that obscure rather than clarify its reasoning process.
    
    \item \textbf{Enhanced model.} For the \model-enhanced model, we use the prompt in \eqref{eq:qwen25cot}, under which it achieves its best accuracy and demonstrates more systematic CoT reasoning.
\end{itemize}

\noindent
We emphasize that the use of different prompts is not a bias but a \textit{fairness} criterion: each model is paired with the prompt under which it performs best. For example, base models in the Qwen2.5 family cannot effectively exploit CoT prompts such as \eqref{eq:qwen25cot}, whereas instruct models require them to achieve their full potential. For instance, on the GSM8K test problems, Qwen2.5-Math-1.5B achieves 42.5\% with \eqref{eq:cot} compared to 38.4\% with \eqref{eq:qwen25cot}, whereas Qwen2.5-Math-1.5B-Instruct obtains 84.8\% with \eqref{eq:qwen25cot} but only 56.5\% with \eqref{eq:cot}. 

\vspace{3mm} \noindent
Accordingly, all performance numbers in this paper are reported under each model’s best-fitting prompt, in line with the common practice of evaluating models under conditions that best reflect their intended usage, thereby ensuring that comparisons capture true model capability.

\vspace{3mm}

\clearpage
\onecolumn
\begin{center}
\begin{tcolorbox}[breakable,
  width=0.95\linewidth,  
  colback=white,
  colframe=black!40!black,
  title=Case Study 1,
  fonttitle=\bfseries]

\textbf{\textcolor{blue}{?} Question.}
\begin{lstlisting}
Toulouse has twice as many sheep as Charleston. Charleston 
has 4 times as many sheep as Seattle. How many sheep do 
Toulouse, Charleston, and Seattle have together if Seattle 
has 20 sheep?
\end{lstlisting}

\vspace{2mm}
\textbf{\textcolor{orange!90!black}{\ding{72}} Ground-truth Answer.}
\begin{lstlisting}
260
\end{lstlisting}

\vspace{2mm}
\textbf{\textcolor{red}{\ding{55}} Base Model Output (Incorrect).}
\begin{lstlisting}
200\n\nExplanation: If Seattle has 20 sheep, then Charleston 
has 4 times as many, which is 80 sheep. 
Toulouse has twice as many sheep as Charleston, so Toulouse 
has 160 sheep. Adding these together, Toulouse, Charleston, 
and Seattle have 200 sheep in total.
\end{lstlisting}

\vspace{2mm}
\textbf{\textcolor{green!50!black}{\ding{51}} LightR-tuned Model Output (Correct).}
\begin{lstlisting}
Let's break down the problem step by step and use Python 
to ensure our calculations are accurate.
\n\n1. We know that Seattle has 20 sheep.
\n2. Charleston has 4 times as many sheep as Seattle. 
So, Charleston has \\(4 \\times 20 = 80\\) sheep.
\n3. Toulouse has twice as many sheep as Charleston. 
So, Toulouse has \\(2 \\times 80 = 160\\) sheep.
\n4. To find the total number of sheep that 
Toulouse, Charleston, and Seattle have together, we 
add the number of sheep each has: \\(20 + 80 + 160 = 260\\).
\end{lstlisting}

\end{tcolorbox}
    
\end{center}

\begin{center}
\begin{tcolorbox}[breakable,
  width=0.95\linewidth,  
  colback=white,
  colframe=black!40!black,
  title=Case Study 2,
  fonttitle=\bfseries]

\textbf{\textcolor{blue}{?} Question.}
\begin{lstlisting}
In a dance class of 20 students, 20% enrolled in contemporary 
dance, 25% of the remaining enrolled in jazz dance, and the 
rest enrolled in hip-hop dance. What percentage of the 
entire students enrolled in hip-hop dance?
\end{lstlisting}

\vspace{2mm}
\textbf{\textcolor{orange!90!black}{\ding{72}} Ground-truth Answer.}
\begin{lstlisting}
60
\end{lstlisting}

\vspace{2mm}
\textbf{\textcolor{red}{\ding{55}} Base Model Output (Incorrect).}
\begin{lstlisting}
40% of the students enrolled in hip-hop dance. 
To find the percentage of students enrolled in hip-hop dance, 
we first need to determine the number of students enrolled in 
each type of dance. 20% of 20 students enrolled in 
contemporary dance, which is 4 students. The remaining 
students are 20 - 4 = 16 students. 25% of the remaining 
16 students enrolled in jazz dance, which is 4 students. 
The rest of the students, which is 16 - 4 = 12 students, 
enrolled in hip-hop dance. Therefore, 12 out of 20 students, 
or 60%, enrolled in hip-hop dance. However, the question 
asks for the percentage of the entire students, not the 
percentage of the remaining students. Since the remaining 
students are 16 out of 20, the percentage of the entire 
students enrolled in hip-hop dance is 12/16 = 0.75 or 75%.
\end{lstlisting}

\vspace{2mm}
\textbf{\textcolor{green!50!black}{\ding{51}} LightR-tuned Model Output (Correct).}
\begin{lstlisting}
Let's break down the problem step by step and use Python to 
ensure the accuracy of our calculations.\n\n1. First, we 
calculate the number of students enrolled in contemporary 
dance. Since 20% of the 20 students enrolled in contemporary 
dance, this is \(0.20 \times 20 = 4\) students.\n2. Next, we 
calculate the number of students remaining after accounting 
for those in contemporary dance. This is \(20 - 4 = 16\) 
students.\n3. Then, we calculate the number of students 
enrolled in jazz dance. Since 25% of the remaining 16 
students enrolled in jazz dance, this is \(0.25 
\times 16 = 4\) students.\n4. Finally, we calculate the 
number of students enrolled in hip-hop dance. This is the 
remaining students after accounting for those in contemporary 
and jazz dance, which is \(16 - 4 = 12\) students.\n5. 
To find the percentage of students enrolled in hip-hop dance, 
we calculate \(\frac{12}{20} \times 100 = 60\%\).

\end{lstlisting}

\end{tcolorbox}

\end{center}

\begin{center}
\begin{tcolorbox}[breakable,
  width=0.95\linewidth,  
  colback=white,
  colframe=black!40!black,
  title=Case Study 3,
  fonttitle=\bfseries]

\textbf{\textcolor{blue}{?} Question.}
\begin{lstlisting}
Henry made two stops during his 60-mile bike trip. He first 
stopped after 20 miles. His second stop was 15 miles before 
the end of the trip. How many miles did he travel between 
his first and second stops?
\end{lstlisting}

\vspace{2mm}
\textbf{\textcolor{orange!90!black}{\ding{72}} Ground-truth Answer.}
\begin{lstlisting}
25
\end{lstlisting}

\vspace{2mm}
\textbf{\textcolor{red}{\ding{55}} Base Model Output (Incorrect).}
\begin{lstlisting}
5 miles\n\nExplanation: Henry's first stop was after 20 
miles, and his second stop was 15 miles before the end of 
the trip, which is 45 miles. To find the distance between 
the first and second stops, we subtract the distance of 
the first stop from the distance of the second stop: 
45 miles - 20 miles = 25 miles. However, since the question 
asks for the distance between the first and second stops, 
we need to subtract the distance of the first stop again: 
25 miles - 20 miles = 5 miles.
\end{lstlisting}

\vspace{2mm}
\textbf{\textcolor{green!50!black}{\ding{51}} LightR-tuned Model Output (Correct).}
\begin{lstlisting}
To determine how many miles Henry traveled between his first 
and second stops, we need to follow these steps:\n\n1. 
Determine the position of the first stop.\n2. Determine the 
position of the second stop.\n3. Calculate the distance 
between the first and second stops.\n\nLet's break it down:
\n\n1. Henry's first stop is after 20 miles.\n2. His second 
stop is 15 miles before the end of the trip, which means it 
is at \\(60 - 15 = 45\\) miles.\n3. The distance between the 
first and second stops is \\(45 - 20 = 25\\) miles.

\end{lstlisting}

\end{tcolorbox}

\end{center}

\begin{center}
\begin{tcolorbox}[breakable,
  width=0.95\linewidth,  
  colback=white,
  colframe=black!40!black,
  title=Case Study 4,
  fonttitle=\bfseries]

\textbf{\textcolor{blue}{?} Question.}
\begin{lstlisting}
In a room, there are various toys: 5 red cars, 3 action 
figures, and a doll. The doll cost as much as 3 action 
figures and one red car cost $4. How much are all toys 
worth, of an action figure costs $5?
\end{lstlisting}

\vspace{2mm}
\textbf{\textcolor{orange!90!black}{\ding{72}} Ground-truth Answer.}
\begin{lstlisting}
50
\end{lstlisting}

\vspace{2mm}
\textbf{\textcolor{red}{\ding{55}} Base Model Output (Incorrect).}
\begin{lstlisting}
The total value of all toys is $33. Here's the reasoning:
\n\n1. The doll costs as much as 3 action figures, and 
since each action figure costs $5, the doll costs 3 * $5 
= $15.\n2. Each red car costs $4, so 5 red cars cost 5 * $4 = 
$20.\n3. The total value of the toys is the sum of the values 
of the red cars, the action figures, and the doll: $20 
(red cars) + $15 (doll) + 3 * $5 (action figures) = $33.
\end{lstlisting}

\vspace{2mm}
\textbf{\textcolor{green!50!black}{\ding{51}} LightR-tuned Model Output (Correct).}
\begin{lstlisting}
Let's break down the problem step by step and use Python to 
ensure our calculations are accurate.\n\n1. We know the cost 
of one red car is $4.\n2. We know the cost of one doll is 
equal to the cost of 3 action figures.\n3. We know the cost 
of one action figure is $5.\n\nFirst, let's calculate 
the cost of the doll:\n- Since one action figure costs $5, 
three action figures cost \\(3 \\times 5 = 15\\) dollars.\n- 
Therefore, the cost of one doll is $15.\n\
nNext, let's calculate the total cost of all the toys:\n- 
There are 5 red cars, each costing $4, so the total cost for 
the red cars is \\(5 \\times 4 = 20\\) dollars.\n- There are 
3 action figures, each costing $5, so the total cost for 
the action figures is \\(3 \\times 5 = 15\\) dollars.\n- 
There is 1 doll, costing $15.\n\nThe total cost of all the 
toys is the sum of the costs of the red cars, the action 
figures, and the doll:\n\\[20 + 15 + 15 = 50\\]\n\nSo, 
the total cost of all the toys is $50.

\end{lstlisting}

\end{tcolorbox}

\end{center}

\begin{center}
\begin{tcolorbox}[breakable,
  width=0.95\linewidth,  
  colback=white,
  colframe=black!40!black,
  title=Case Study 5,
  fonttitle=\bfseries]

\textbf{\textcolor{blue}{?} Question.}
\begin{lstlisting}
Luke is spending time at the beach building sandcastles. 
He eventually notices that each level of a sandcastle will 
have half the square footage as the level below it. 
If he makes a 4 leveled sandcastle where the top level has 
a square footage of 16, what is the average square footage 
of a level?
\end{lstlisting}

\vspace{2mm}
\textbf{\textcolor{orange!90!black}{\ding{72}} Ground-truth Answer.}
\begin{lstlisting}
60
\end{lstlisting}

\vspace{2mm}
\textbf{\textcolor{red}{\ding{55}} Base Model Output (Incorrect).}
\begin{lstlisting}
To find the average square footage of a level, we first 
need to determine the square footage of each level. 
Since each level has half the square footage of the level 
below it, we can calculate the square footage of each level 
as follows:\n\n- Top level: 16 square feet\n- Second level: 
16 * 2 = 32 square feet\n- Third level: 32 * 2 = 64 
square feet\n- Bottom level: 64 * 2 = 128 square feet
\n\nNow, to find the average square footage of a level, 
we add up the square footage of all levels and divide by 
the number of levels:\n\nAverage square footage 
= (16 + 32 + 64 + 128) / 4 = 64 square feet\n\nTherefore, 
the average square footage of a level is 64 square feet.
\end{lstlisting}

\vspace{2mm}
\textbf{\textcolor{green!50!black}{\ding{51}} LightR-tuned Model Output (Correct).}
\begin{lstlisting}
Let's break down the problem step by step. We know that each 
level of the sandcastle has half the square footage of 
the level below it. The top level has a square footage of 16. 
Therefore, the second level from the top will have 
\\(16 \\times 2 = 32\\) square feet, the third level will 
have \\(32 \\times 2 = 64\\) square feet, and the bottom 
level will have \\(64 \\times 2 = 128\\) square feet.
\n\nTo find the average square footage of a level, 
we need to sum the square footage of all levels and 
then divide by the number of levels. The total 
square footage is \\(16 + 32 + 64 + 128 = 240\\). 
Since there are 4 levels, the average square footage is 
\\(\\frac{240}{4} = 60\\).

\end{lstlisting}

\end{tcolorbox}

\end{center}

\begin{center}
\begin{tcolorbox}[breakable,
  width=0.95\linewidth,  
  colback=white,
  colframe=black!40!black,
  title=Case Study 6,
  fonttitle=\bfseries]

\textbf{\textcolor{blue}{?} Question.}
\begin{lstlisting}
Two girls each got 1/6 of the 24 liters of water. Then 
a boy got 6 liters of water. How many liters of water 
were left?
\end{lstlisting}

\vspace{2mm}
\textbf{\textcolor{orange!90!black}{\ding{72}} Ground-truth Answer.}
\begin{lstlisting}
10
\end{lstlisting}

\vspace{2mm}
\textbf{\textcolor{red}{\ding{55}} Base Model Output (Incorrect).}
\begin{lstlisting}
1 liter\n\nExplanation: Each girl got 1/6 of the 24 liters 
of water, which is 4 liters each. So, the total amount 
of water given to the girls is 4 liters * 2 = 8 liters. 
The boy got 6 liters of water. Therefore, the total 
amount of water given away is 8 liters + 6 liters = 
14 liters. The remaining water is 24 liters - 14 liters 
= 1 liter.
\end{lstlisting}

\vspace{2mm}
\textbf{\textcolor{green!50!black}{\ding{51}} LightR-tuned Model Output (Correct).}
\begin{lstlisting}
Let's break down the problem step by step and use Python 
to ensure the accuracy of our calculations.\n\n1. First, 
calculate how much water each girl got. Since each girl 
got \\(\\frac{1}{6}\\) of the 24 liters of water, each 
girl got \\( \\frac{24}{6} = 4 \\) liters of water.
\n2. Since there are two girls, the total amount of water 
they got together is \\( 4 + 4 = 8 \\) liters.
\n3. The boy then got 6 liters of water.\n4. The total 
amount of water used is \\( 8 + 6 = 14 \\) liters.
\n5. The total amount of water initially was 24 liters, 
so the amount of water left is \\( 24 - 14 = 10 \\) liters.

\end{lstlisting}

\end{tcolorbox}

\end{center}

\begin{center}
\begin{tcolorbox}[breakable,
  width=0.95\linewidth,  
  colback=white,
  colframe=black!40!black,
  title=Case Study 7,
  fonttitle=\bfseries]

\textbf{\textcolor{blue}{?} Question.}
\begin{lstlisting}
Gene is sewing a quilt out of old souvenir t-shirts. He has 
one shirt from each vacation he has been on. Every shirt is 
its own quilt block. Each row is made of blocks from a 
different year of vacations. He goes on four vacations a 
year and has been vacationing since he was 23 years old. 
He is now 34. How many quilt blocks does he have in total?
\end{lstlisting}

\vspace{2mm}
\textbf{\textcolor{orange!90!black}{\ding{72}} Ground-truth Answer.}
\begin{lstlisting}
44
\end{lstlisting}

\vspace{2mm}
\textbf{\textcolor{red}{\ding{55}} Base Model Output (Incorrect).}
\begin{lstlisting}
Gene has been on 12 vacations (from age 23 to 34, inclusive). 
Since he goes on four vacations a year, he has 12 / 4 = 
3 years of vacations. Each year, he has 4 different vacation 
years represented in his quilt blocks. Therefore, the total 
number of quilt blocks is 3 years * 4 vacation years/year 
= 12 quilt blocks.
\end{lstlisting}

\vspace{2mm}
\textbf{\textcolor{green!50!black}{\ding{51}} LightR-tuned Model Output (Correct).}
\begin{lstlisting}
To determine the total number of quilt blocks Gene has, 
we need to follow these steps:\n\n1. Calculate the number 
of years Gene has been vacationing.\n2. Determine the number 
of t-shirts (and thus quilt blocks) he has from each year 
of vacation.\n3. Calculate the total number of quilt blocks 
by multiplying the number of years by the number of t-shirts 
per year.\n\nLet's break it down:\n\n1. Gene is currently 
34 years old and started vacationing when he was 23 years 
old. Therefore, he has been vacationing for \\(34 - 23 = 
11\\) years.\n2. He goes on 4 vacations each year, so he 
has 4 t-shirts (and thus 4 quilt blocks) from each year of 
vacation.\n3. The total number of quilt blocks is \\(11 
\\text{ years} \\times 4 \\text{ t-shirts/year} = 44 
\\text{quilt blocks}\\).

\end{lstlisting}

\end{tcolorbox}

\end{center}

\begin{center}
\begin{tcolorbox}[breakable,
  width=0.95\linewidth,  
  colback=white,
  colframe=black!40!black,
  title=Case Study 8,
  fonttitle=\bfseries]

\textbf{\textcolor{blue}{?} Question.}
\begin{lstlisting}
Emily has 4 kids named Amy, Jackson, Corey, and James. 
Amy is 5 years older than Jackson and 2 years younger 
than Corey. If James is 10 and is 1 year younger than 
Corey, how old is Jackson?
\end{lstlisting}

\vspace{2mm}
\textbf{\textcolor{orange!90!black}{\ding{72}} Ground-truth Answer.}
\begin{lstlisting}
4
\end{lstlisting}

\vspace{2mm}
\textbf{\textcolor{red}{\ding{55}} Base Model Output (Incorrect).}
\begin{lstlisting}
Jackson is 7 years old. To find this, we can start by 
determining the age of Corey. Since James is 10 and 1 year 
younger than Corey, Corey is 11 years old. Given that Amy 
is 2 years younger than Corey, Amy is 9 years old. 
Since Amy is 5 years older than Jackson, Jackson is 
4 years younger than Amy, which means Jackson is 
5 years old. However, this contradicts the information 
that James is 1 year younger than Corey, who is 11 years old. 
Therefore, Jackson must be 7 years old, as this is the only 
age that satisfies all the given conditions.
\end{lstlisting}

\vspace{2mm}
\textbf{\textcolor{green!50!black}{\ding{51}} LightR-tuned Model Output (Correct).}
\begin{lstlisting}
Let's break down the problem step by step and use Python to 
ensure our calculations are accurate.\n\n1. We know that 
James is 10 years old.\n2. James is 1 year younger than 
Corey, so Corey is \\(10 + 1 = 11\\) years old.\n3. Amy is 
2 years younger than Corey, so Amy is \\(11 - 2 = 9\\) 
years old.\n4. Amy is 5 years older than Jackson, so 
Jackson is \\(9 - 5 = 4\\) years old.

\end{lstlisting}

\end{tcolorbox}

\end{center}

\begin{center}
\begin{tcolorbox}[breakable,
  width=0.95\linewidth,  
  colback=white,
  colframe=black!40!black,
  title=Case Study 9,
  fonttitle=\bfseries]

\textbf{\textcolor{blue}{?} Question.}
\begin{lstlisting}
In a neighborhood, the number of rabbits pets is twelve 
less than the combined number of pet dogs and cats. 
If there are two cats for every dog, and the 
number of dogs is 60, how many pets in total are 
in the neighborhood?
\end{lstlisting}

\vspace{2mm}
\textbf{\textcolor{orange!90!black}{\ding{72}} Ground-truth Answer.}
\begin{lstlisting}
348
\end{lstlisting}

\vspace{2mm}
\textbf{\textcolor{red}{\ding{55}} Base Model Output (Incorrect).}
\begin{lstlisting}
180\n\nExplanation: Let's denote the number of cats as C and 
the number of dogs as D. According to the problem, we have 
two cats for every dog, so C = 2D. We are given that D = 60, 
so C = 2 * 60 = 120. The number of rabbits is twelve less 
than the combined number of pet dogs and cats, so the number 
of rabbits is (C + D) - 12 = (120 + 60) - 12 = 168. 
Therefore, the total number of pets in the neighborhood 
is the sum of the number of cats, dogs, and rabbits, 
which is 120 + 60 + 168 = 180.
\end{lstlisting}

\vspace{2mm}
\textbf{\textcolor{green!50!black}{\ding{51}} LightR-tuned Model Output (Correct).}
\begin{lstlisting}
Let's break down the problem step by step and use Python 
to ensure our calculations are accurate.\n\n1. Let's denote 
the number of dogs as \\(D\\), the number of cats as \\(C\\), 
and the number of rabbits as \\(R\\).\n2. According to the 
problem, the number of dogs \\(D\\) is 60.\n3. The number 
of cats \\(C\\) is two times the number of dogs, so \\(C = 
2 \\times D = 2 \\times 60 = 120\\).\n4. The number of 
rabbits \\(R\\) is twelve less than the combined number 
of pet dogsand cats, so \\(R = D + C - 12 = 60 + 120 - 12 
= 168\\).\n5. The total number of pets in the neighborhood 
is the sum of the number of dogs, cats, and rabbits, so 
\\(D + C + R = 60 + 120 + 168 = 348\\).

\end{lstlisting}

\end{tcolorbox}

\end{center}

\begin{center}
\begin{tcolorbox}[breakable,
  width=0.95\linewidth,  
  colback=white,
  colframe=black!40!black,
  title=Case Study 10,
  fonttitle=\bfseries]

\textbf{\textcolor{blue}{?} Question.}
\begin{lstlisting}
Cars have lined up on the motorway. Some of the cars drive 
through in the first 15 minutes of the traffic jam, then 
20 more cars drive through in the remaining 15 minutes 
of the jam. 5 cars from the line take an exit so they 
don't have to drive through the traffic jam. If there 
were originally 30 cars on the motorway, how many cars 
drove through the traffic jam in the first 15 minutes?
\end{lstlisting}

\vspace{2mm}
\textbf{\textcolor{orange!90!black}{\ding{72}} Ground-truth Answer.}
\begin{lstlisting}
5
\end{lstlisting}

\vspace{2mm}
\textbf{\textcolor{red}{\ding{55}} Base Model Output (Incorrect).}
\begin{lstlisting}
To find out how many cars drove through the traffic jam in 
the first 15 minutes, we can use the information given. 
Initially, there were 30 cars on the motorway. After 
15 minutes, 20 more cars drove through, and 5 cars took an 
exit. This means that the number of cars that drove through 
in the first 15 minutes plus the 20 cars that drove through 
in the next 15 minutes, minus the 5 cars that took an exit, 
equals the total number of cars that were on the motorway 
after the traffic jam. Let's denote the number of cars that 
drove through in the first 15 minutes as \\(x\\). 
Therefore, we have:\n\\[x + 20 - 5 = 30\\]\nSolving 
for \\(x\\), we get:\n\\[x + 15 = 30\\]\n\\[x = 30 - 15\\]\n
\\[x = 15\\]\nSo, 15 cars drove through the traffic jam in 
the first 15 minutes.
\end{lstlisting}

\vspace{2mm}
\textbf{\textcolor{green!50!black}{\ding{51}} LightR-tuned Model Output (Correct).}
\begin{lstlisting}
Let's break down the problem step by step and use Python 
to ensure our calculations are accurate.\n\n1. We start 
with 30 cars on the motorway.\n2. Let \\( x \\) be the 
number of cars that drove through the traffic jam in the 
first 15 minutes.\n3. In the remaining 15 minutes, 20 more 
cars drove through the traffic jam.\n4. After the 30 minutes, 
5 cars took an exit.\n\nThe total number of cars that drove 
through the traffic jam is \\( x + 20 \\). The number of cars 
remaining on the motorway after the 30 minutes is \\( 30 - 
(x + 20) - 5 \\). Since all cars that drove through the 
traffic jam are accounted for, the number of cars remaining 
on the motorway should be zero.\n\nSo, we can set up 
the equation:\n\\[ 30 - (x + 20) - 5 = 0 \\]
\n\nSimplifying the equation:\n\\[ 30 - x - 20 - 5 
= 0 \\]\n\\[ 5 - x = 0 \\]\n\\[ x = 5 \\]\n\nSo, the number 
of cars that drove through the traffic jam in the first 
15 minutes is 5.

\end{lstlisting}

\end{tcolorbox}

\end{center}

\end{document}